\documentclass[letterpaper, 10 pt, journal, twoside]{ieeetran}

\IEEEoverridecommandlockouts                              

\usepackage{xcolor}
\usepackage{amsmath}
\usepackage{graphicx}
\usepackage{subcaption}
\usepackage{float}
\usepackage{algorithm}
\usepackage{algpseudocode}
\usepackage{hyperref}
\DeclareMathOperator*{\argmax}{arg\,max}

\usepackage{tikz}
\usepackage{textcomp}
\usepackage{hyperref}
\usepackage{lipsum}

\newcommand\copyrighttext{%
  \footnotesize \textcopyright~ 2020 IEEE. Personal use of this material is permitted.  Permission from IEEE must be obtained for all other uses, in any current or future media, including reprinting/republishing this material for advertising or promotional purposes, creating new collective works, for resale or redistribution to servers or lists, or reuse of any copyrighted component of this work in other works.}
\newcommand\copyrightnotice{%
\begin{tikzpicture}[remember picture,overlay]
\node[anchor=south,yshift=5pt] at (current page.south) {\fbox{\parbox{\dimexpr\textwidth-\fboxsep-\fboxrule\relax}{\copyrighttext}}};
\end{tikzpicture}%
}

\graphicspath{{./figs/}{./figs/testcases/}}

\title{
A Deep Learning Approach to Grasping the Invisible
}

\author{Yang Yang$^{1}$, Hengyue Liang$^{2}$ and Changhyun Choi$^{2}$
\thanks{Manuscript received: September, 10, 2019; Revised December, 21, 2019; Accepted January, 14, 2020.}
\thanks{This paper was recommended for publication by Editor Hong Liu upon evaluation of the Associate Editor and Reviewers' comments.
This work was in part supported by the MnDRIVE Initiative on Robotics, Sensors, and Advanced Manufacturing.} 
\thanks{$^{1}$Y. Yang is with the Department of Computer Science and Engineering, Univ. of Minnesota, Minneapolis, USA {\tt\small yang5276@umn.edu}}%
\thanks{$^{2}$ H. Liang and C. Choi are with the Department of Electrical and Computer Engineering, Univ. of Minnesota, Minneapolis, USA {\tt\small \{liang656, cchoi\}@umn.edu}}%
\thanks{Digital Object Identifier (DOI): see top of this page.} 
}

\begin{document}

\markboth{IEEE Robotics and Automation Letters. Preprint Version. Accepted January, 2020}
{Yang \MakeLowercase{\textit{et al.}}: A Deep Learning Approach to Grasping the Invisible} 

\maketitle

\copyrightnotice

\begin{abstract}

We study an emerging problem named ``grasping the invisible'' in robotic manipulation, in which a robot is tasked to grasp an initially invisible target object via a sequence of pushing and grasping actions. In this problem, pushes are needed to search for the target and rearrange cluttered objects around it to enable effective grasps. We propose to solve the problem by formulating a deep learning approach in a critic-policy format. The target-oriented motion critic, which maps both visual observations and target information to the expected future rewards of pushing and grasping motion primitives, is learned via deep Q-learning. We divide the problem into two subtasks, and two policies are proposed to tackle each of them, by combining the critic predictions and relevant domain knowledge. A Bayesian-based policy accounting for past action experience performs pushing to search for the target; once the target is found, a classifier-based policy coordinates target-oriented pushing and grasping to grasp the target in clutter. The motion critic and the classifier are trained in a self-supervised manner through robot-environment interactions. Our system achieves a 93\% and 87\% task success rate on each of the two subtasks in simulation and an 85\% task success rate in real robot experiments on the whole problem, which outperforms several baselines by large margins. Supplementary material is available at \href{https://sites.google.com/umn.edu/grasping-invisible}{https://sites.google.com/umn.edu/grasping-invisible}.

\end{abstract}

\begin{IEEEkeywords}
Dexterous Manipulation, Deep Learning in Robotics and Automation, Computer Vision for Automation
\end{IEEEkeywords}

\section{INTRODUCTION}\label{sec:introduction}

\begin{figure}[t]
    \centering
    \begin{subfigure}{0.13\textwidth}
        \centering
        \includegraphics[scale=0.31]{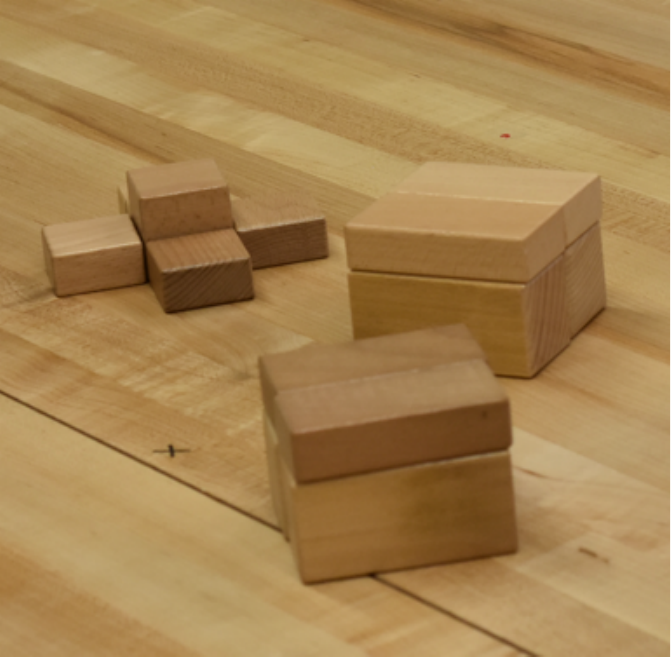}%
        \caption{Configuration}
        \label{fig:bury}
    \end{subfigure}
    \hfill
    \begin{subfigure}{0.34\textwidth}  
        \centering
        \includegraphics[scale=0.31]{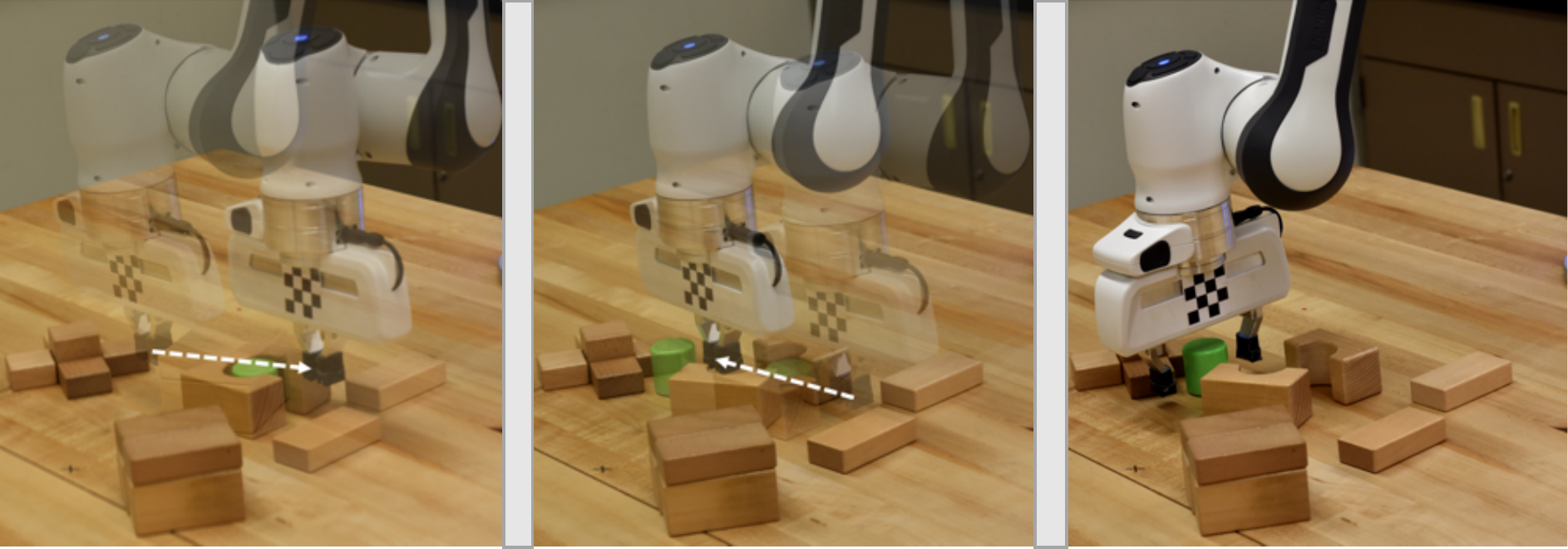}%
        \caption{Our approach}
    \end{subfigure}
    \vspace{-3pt}
    \caption{\textbf{Example configuration of the ``grasping the invisible'' problem}. The target object is the green cylinder and initially invisible to the robot. We propose to solve the problem with an exploration-singulation-grasping procedure.}
    \vspace{-3pt}
    \label{fig:example}
\end{figure}

\IEEEPARstart{I}{magine} what happens when a young child is looking for a specific toy block buried in clutter, as shown in Fig. \ref{fig:bury}. He/she may first push apart the pile of blocks and luckily spot the target block in clutter, then push around the target block to make space for the fingers (we refer to this type of motion as ``singulation'' \cite{eitel17isrr}) and finally grasp it. We wonder if an intelligent agent can perform such a task. To grasp an invisible target, a robot is required to have the ability of explorational pushing, singulation by target-oriented pushing (i.e., push the target or surrounding objects to make some space), and target-oriented grasping.

Robot pushing \cite{eitel17isrr}, grasping \cite{choi2018learning} or push-grasping \cite{zeng2018learning} has been actively studied but mostly on relatively simple target-agnostic tasks. Without incorporating target information effectively, fast adaptations of these methods (e.g., by applying the target mask directly on the results of the methods) for target-oriented tasks are not successful. While some target-oriented manipulation approaches \cite{jang2018grasp2vec}, \cite{fang2018multi} have been applied to the visible target in sparse environments (i.e., the target is well isolated), the scenarios to apply these methods are limited.

Moving forward from target-agnostic robotic manipulation to target-oriented manipulation challenges not only perception but also manipulation capabilities. Specifically, these challenges are 1) the target information is missing in the case of complete occlusion, and efficient explorations are thus required, 2) perception modules responsible for reasoning about the target region work poorly in dense clutter, 3) data labeling in target-oriented tasks is expensive and 4) the coordination between pushing and grasping (i.e., deciding when and how to push or grasp) is critical but tricky to accomplish.

In this paper, we propose to solve the ``grasping the invisible'' problem by formulating a deep learning approach in a critic-policy format. The key aspects of our system are
\begin{itemize}
    \item A robust semantic segmentation module is used to annotate the objects of interest and detect the existence of the target.
    \item We learn a target-oriented motion critic through deep Q-learning. The critic takes as input visual observations and the target mask and predicts expected future rewards (i.e., Q values) for target-oriented pushing and grasping motion primitives.
    \item By incorporating Q predictions with domain knowledge, two policies are proposed to make pushing or grasping action decisions in different scenarios. Specifically, a Bayesian-based policy accounting for action experience performs efficient explorational pushing in the complete occlusion. Once the target is visible, coordinated decision making in target-oriented pushing or grasping is achieved by a classifier-based policy that takes the clutteredness around the target as an input.
    \item Our learning models (the critic and the classifier) are fully self-supervised through robot-environment interactions.
\end{itemize}
Fig. \ref{fig:example} presents an example configuration of the ``grasping the invisible'' problem and how we propose to solve it.

\textbf{Contributions} This paper presents two core technical contributions: 1) a motion critic for target-oriented pushing and grasping motion primitives and 2) a Bayesian-based policy for target exploration and a classifier-based policy for coordinated decision making. Our system can perform target-oriented manipulation tasks with observations from an RGB-D camera.
\begin{figure*}[t]
  \centering
  \includegraphics[width=\textwidth]{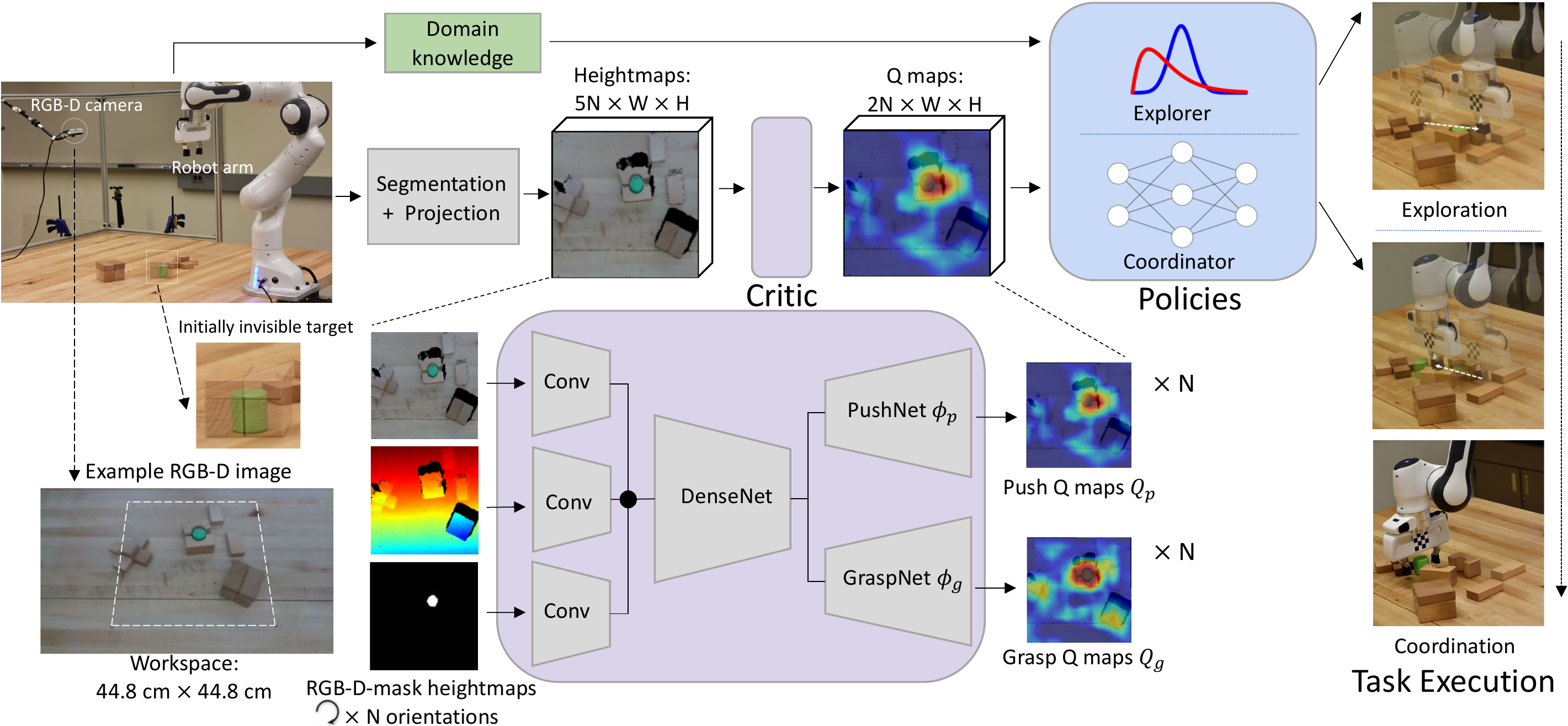}
  \vspace{-10pt}
  \caption{\textbf{Overview.} The visual observations of the scene (we use an example image containing a visible target for illustration) and the target mask from semantic segmentation are orthographically projected to construct heightmaps. The heightmaps are rotated by $N$ orientations for different motion angles and then fed into the motion critic. The critic predicts pixel-wise push and grasp Q values. The two policies, explorer and coordinator, take as input Q predictions and specific domain knowledge to search for the invisible target (exploration) and coordinate pushing and grasping (coordination).}
  \vspace{-5pt}
  \label{fig:overview}
\end{figure*}
\section{RELATED WORK}
Robotic grasping has been well and successfully studied. Classic model-driven approaches \cite{bohg2013data}, \cite{sahbani2012overview} find stable force closures for known objects by utilizing prior knowledge such as 3D models of manipulators and objects and their physical properties. Recent data-driven approaches \cite{mahler2017dex}, \cite{kalashnikov2018scalable}, \cite{choi2018learning} harness learning algorithms and data (collected from humans or physical experiments) to directly map visual observations to grasp representations. Our approach is data-driven and model-agnostic, and the learning models are self-supervised. 

To mitigate uncertainty and collision introduced by clutter, non-prehensile manipulation \cite{dogar2012planning}, such as pushing, are investigated in both model-driven \cite{cosgun2011push}, \cite{hermans2012guided}, \cite{chang2012interactive} and data-driven approaches \cite{eitel17isrr}, \cite{danielczuk2018linear}, \cite{marios2019robust}. With the addition of pushing, push-grasping systems \cite{boularias2015learning}, \cite{zeng2018learning} are advanced. Analogous to these methods, our approach utilizes non-prehensile pushing to facilitate grasping but further considers both target exploration and singulation.

In comparison to target-agnostic grasping discussed above, target-oriented grasping has been much less explored, except for \cite{jang2017end}, \cite{jang2018grasp2vec}, \cite{fang2018multi}, \cite{marios2019robust}. In \cite{jang2018grasp2vec}, object representations are learned via autonomous robot interaction with the environment for target-oriented grasping. The work in \cite{fang2018multi} presents a multi-task domain adaptation framework to transfer the learned target-oriented grasping policy from simulation to the real world. The environments in these works, however, tend to be sparse, and thus the application scenarios are heavily limited. Moreover, visibility of the target is a strict precondition for representation computation in \cite{jang2018grasp2vec} or mask extraction in \cite{fang2018multi}. In contrast, our method does not assume initial visibility of the target and takes advantage of target-oriented pushing to grasp the target in challenging clutter. \cite{marios2019robust} employs a target-oriented pushing policy analogous to ours, in which a sole push policy is learned via Q-learning for the visible target in clutter. In contrast, we utilize both pushing and grasping for the (visible or invisible) target, and thus the application scenarios are enlarged. In \cite{marios2019robust}, the singulation of the target is confirmed by a hand-crafted module which checks the minimum distance from the nearest object. However, we employ a neural network that learns to decide whether to push or grasp from self-experience. This implicit singulation scheme tends to improve the action efficiency since complete isolation is not necessary for a successful grasping in clutter, as shown in Fig. \ref{fig:rewards}.

One method that motivates our study is visual pushing for grasping (VPG) by Zeng et al. \cite{zeng2018learning}.  VPG proposes a Q-learning framework to learn complementary pushing and grasping policies for robot picking tasks. In VPG, the robot performs target-agnostic tasks and removes all objects from the workspace. In contrast, our approach learns the critic for target-oriented manipulation and proposes subtask policies to solve a more general and complex problem, ``grasping the invisible''.

A recent work by Danielczuk et al. \cite{danielczuk2019mechanical} proposes action heuristics to choose between grasping, suction, and pushing actions to retrieve a target occluded in clutter. (See Appendix \ref{appen:mech_search} for more details.) The main differences between our work and \cite{danielczuk2019mechanical} are two-fold: 1) Problem formulation. \cite{danielczuk2019mechanical} formulates the problem as a POMDP but solves it by hard-coded heuristics. In contrast, we learn a target-oriented motion critic with an MDP formulation when the target is visible. The critic is further utilized in both visible and invisible cases by two policies. 2) Utilization of pushing. In \cite{danielczuk2019mechanical}, pushing is given a lower priority by the heuristics, and makes up only $5\%$ of executed actions. Instead, we use a learned classifier to coordinate pushing and grasping for solving more challenging arrangements.

\section{PROBLEM FORMULATION}

The ``grasping the invisible'' problem in this paper is formulated as follows:

\vspace{5pt}\textbf{Definition 1.} \textit{Given a description (e.g., the class name) of the target object, the goal is to grasp the target via a finite sequence of pushes and grasps. The target can be placed with arbitrary pose and occlusions in dense clutter.}\vspace{5pt}

To tackle the challenge from various poses and occlusions, we divide the problem into two subtasks:

\vspace{5pt}\textbf{Subtask 1.} \textit{If the target is completely buried in clutter, then the robot searches for the target and breaks the structure to make it visible. We name this the \textbf{exploration} task.}

\vspace{5pt}\textbf{Subtask 2.} \textit{Though clearly visible, the target might be closely surrounded by other objects, leaving no space for grasping. Thus sole target-oriented grasping is impossible or inefficient without breaking the structured clutter by target-oriented pushing. In the \textbf{coordination} task, the target-oriented pushing and grasping need to be sequentially coordinated to grasp the target with the most action efficiency.}

\section{Making Target-oriented action decision}
\subsection{System Overview}
As shown in Fig. \ref{fig:overview}, a fixed-mount RGB-D camera captures the predefined workspace. The RGB image is first passed into a pre-trained semantic segmentation module to predict a target mask. The segmentation module robustly detects the target mask, even in heavy occlusions. (See Appendix \ref{appen:seg} for more details.) Then RGB, depth, and mask images are orthographically projected in the gravity direction with a known extrinsic parameter of the camera to construct color heightmap $c_t$, depth heightmap $d_t$, and target mask heightmap $m_t$. 

The motion critic takes as input the heightmaps, and the mask heightmap specifies the target. Under the assumption of a visible target, the critic predicts pixel-wise critic scores (i.e., Q values) for target-oriented pushing and grasping motion primitives. And we find suitable action execution location and rotation based on the critic predictions. We discuss the motion critic in detail in Sec. \ref{sec:critic}.

Every pixel in the Q maps parameterizes a primitive pushing or grasping action, so there is a direct mapping from Q maps to primitive motions. Every 2D pixel is mapped to a 3D action execution position through the depth heightmap with a heuristic distance. Different motion angles are achieved by rotating the input heightmaps by $N$ orientations before the heightmaps are fed into the networks. There are $N$ corresponding Q maps for pushing and grasping respectively \cite{zeng2018learning}. We choose $N=16$ in our system, and the angle discretion is thus $22.5^\circ$.

To grasp an initially invisible target, top-level policies combine Q predictions and domain knowledge for task execution. First, an exploration policy (explorer) decides a pushing location to search for the invisible target. The explorer uses the height distribution of the workspace and the history of previous actions as the domain knowledge. Once the target is found, a coordination policy (coordinator) coordinates target-oriented pushing and grasping by considering the clutteredness around the target as domain knowledge. We discuss the formulation of the two policies in the following Sec. \ref{sub:policy}.

\subsection{Policy} \label{sub:policy}
\begin{figure}[b]
    \centering
    \begin{subfigure}{0.11\textwidth}
        \centering
        \includegraphics[width=\textwidth]{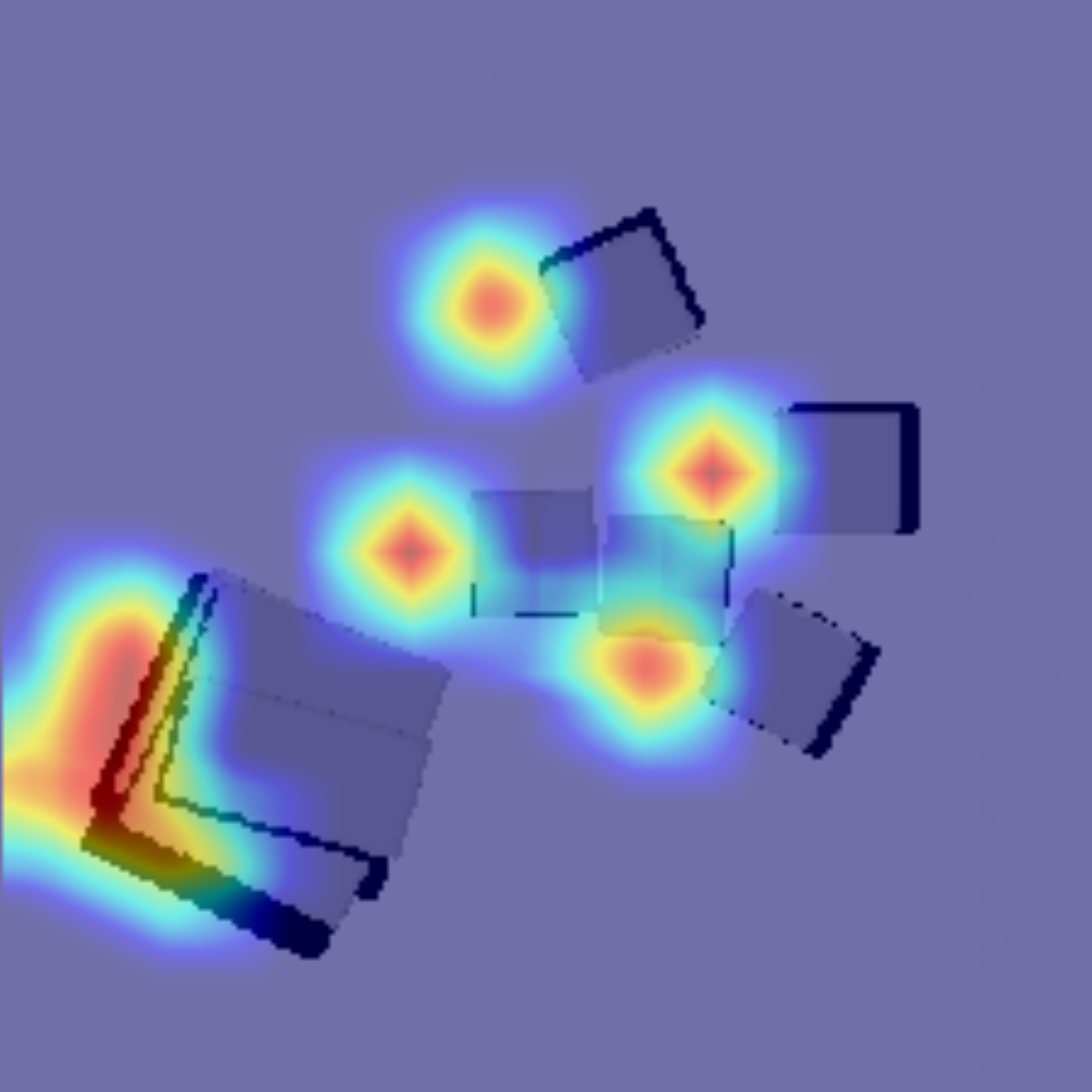}
        \caption*{$C_p$}
    \end{subfigure}
    \begin{subfigure}{0.11\textwidth}
        \centering
        \includegraphics[width=\textwidth]{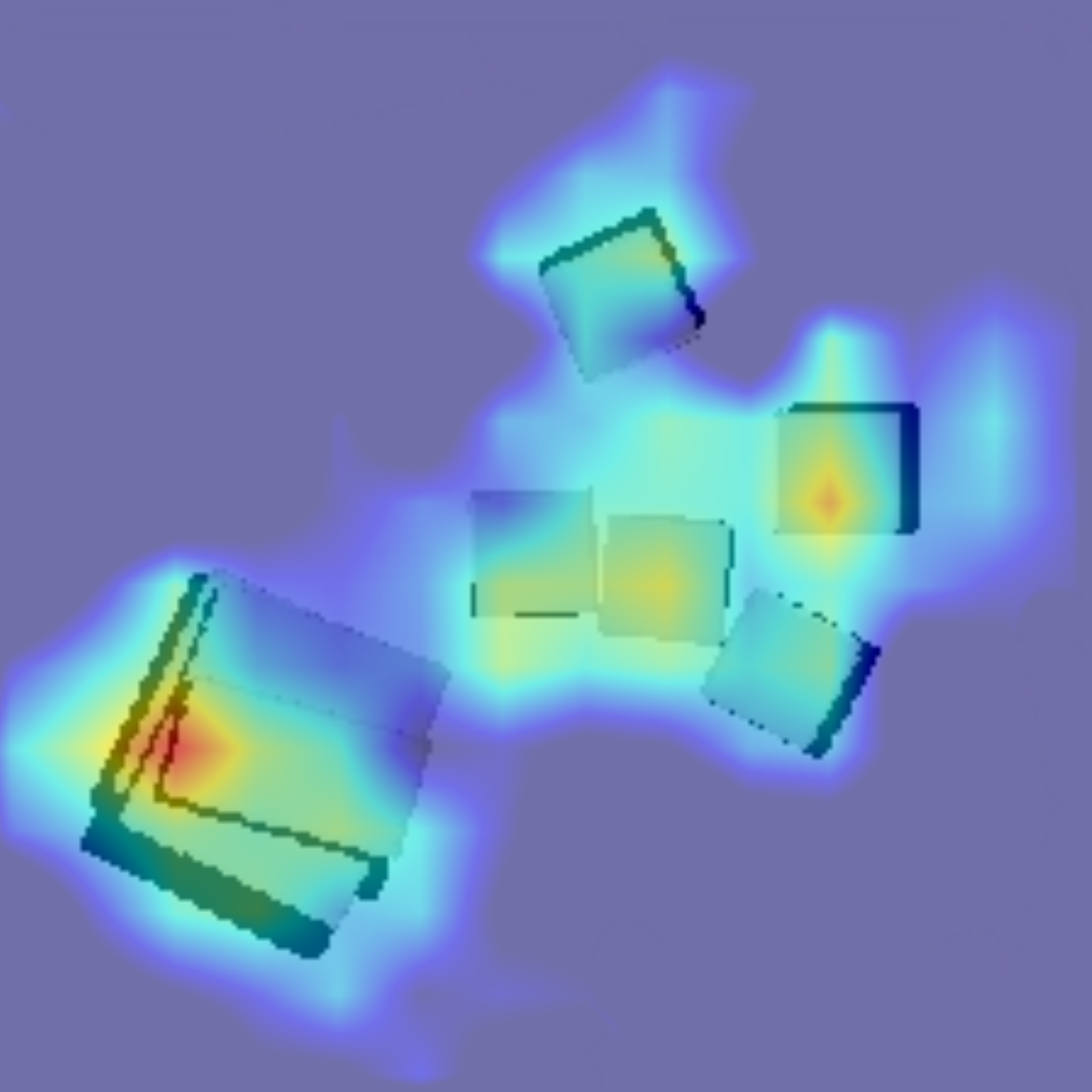}
        \caption*{$A_p$}
    \end{subfigure}
    \begin{subfigure}{0.11\textwidth}
        \centering
        \includegraphics[width=\textwidth]{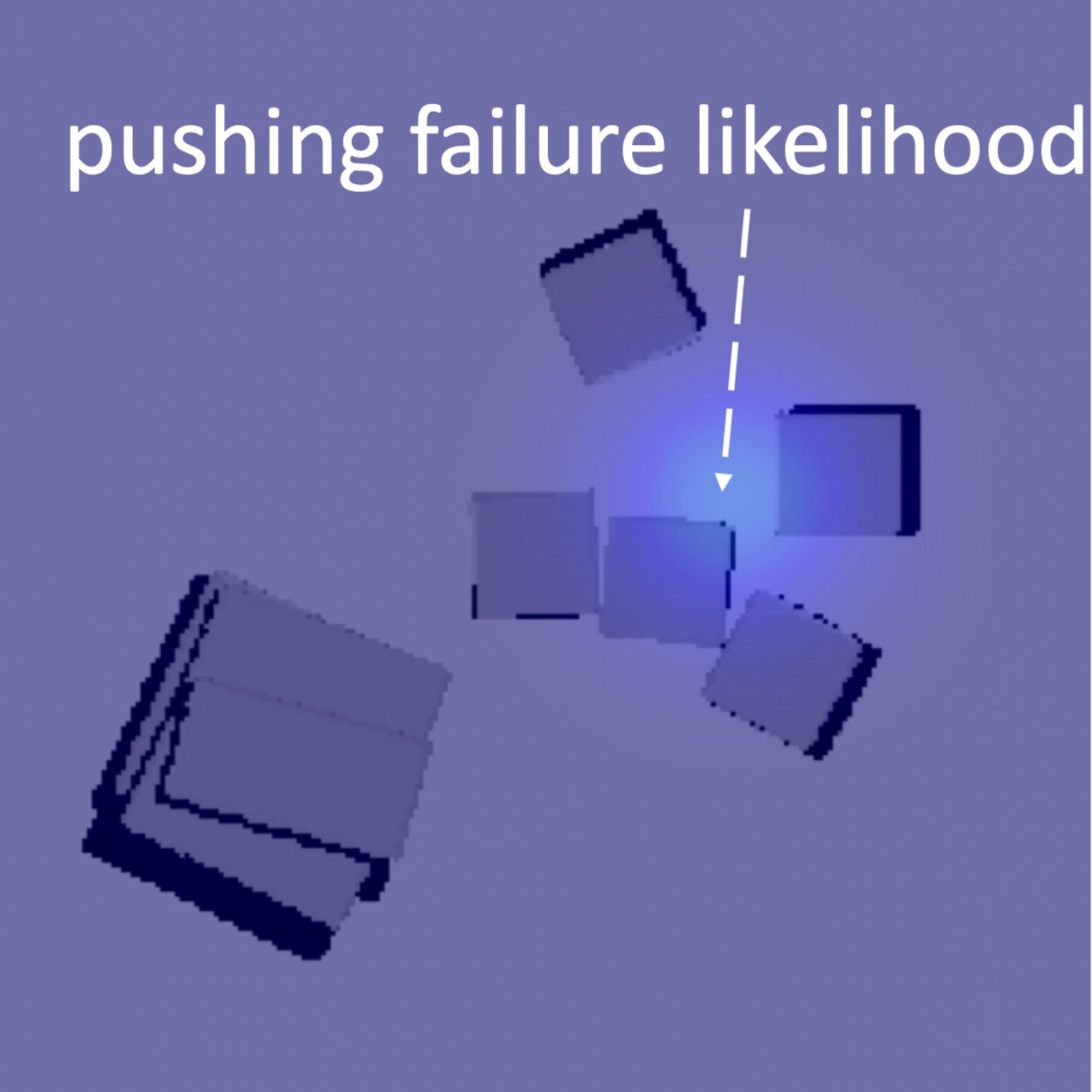}
        \caption*{$F_p$}
    \end{subfigure}
    \begin{subfigure}{0.11\textwidth}
        \centering
        \includegraphics[width=\textwidth]{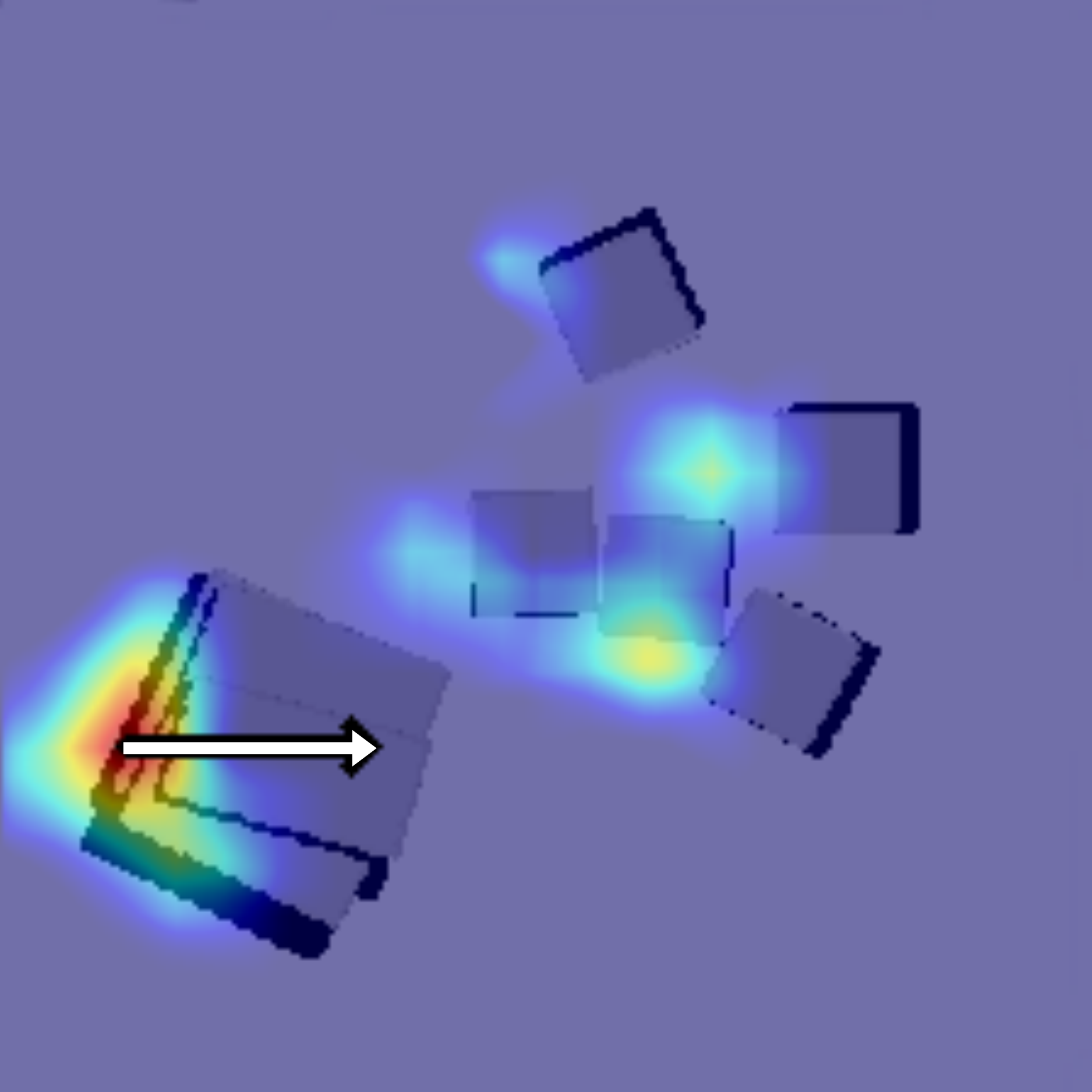}
        \caption*{Posterior}
    \end{subfigure}
    \vspace{-2pt}
    \caption{\textbf{Example of exploration probability maps}. Clutter prior $C_p$, target-agnostic push maps $A_p$ and pushing failure likelihood $F_p$ are multiplied to construct posterior probability maps, according to which the explorational pushing is executed. Only the map representing the intended pushing orientation is visualized here.}
    \vspace{-7pt}
\label{fig:exploration_maps}
\end{figure}

\textbf{Exploration policy (explorer).} To effectively search for the invisible target in the workspace, we propose a Bayesian-based policy $\pi_e$ to push\footnote{Pushing is empirically more effective in our setting, while grasping is inefficient or often impossible in challenging arrangements.}. As shown in Fig. \ref{fig:exploration_maps}, we use the product of target-agnostic push maps $A_p$ and clutter prior $C_p$ as the prior probability for searching. All-ones mask \cite{fang2018multi} is used as an input to the target-oriented push network $\phi_p$, representing all objects in the workspace as the potential target. Thus the corresponding push Q maps $A_p$ give target-agnostic pushing skills. Besides, the prior knowledge of clutter is also considered---the surfaces of clutter are usually not flat. $C_p$ is generated from the depth heightmap by detecting varying heights. It gives the prior knowledge about the edges of clutter along the intended pushing direction, in the form of probability maps. $A_p$ and $C_p$ together constitute our prior for exploration. More details are delineated in Sec. \ref{sub:experiments_exploration}.

To avoid the robot getting stuck at local areas, we account for the past failing experience and construct pushing failure likelihood $F_p$, which is a multimodal Gaussian likelihood function with low peaks centered at the three most recent locations of failed pushes. In Appendix \ref{appen:push_failures} we discuss how to construct $F_p$ in detail. In general, $F_p$ reflects the fact that repeated pushes on the previously failed locations are less likely to find the target object. The explorer then makes a pushing decision based on the posterior probability maps
\begin{align}
    \pi_e : \argmax_a F_p \circ (C_p \circ A_p)
\end{align}
where $\circ$ is the Hadamard product, also known as the entrywise product.

\textbf{Coordination policy (coordinator).} To coordinate pushing and grasping, we propose a classifier-based policy, denoted as $\pi_c$. The binary classifier takes as input domain knowledge, as well as maximum push Q value $q_p$ and maximum grasp Q value $q_g$. Specifically, the domain knowledge is target border occupancy ratio $r_b = \frac{o_b}{\sum m_b}$, target border occupancy norm $n_b=\frac{o_b}{\sum m_t}$ (border occupancy value $o_b$, target border $m_b$, target mask $m_t$ defined in Sec. \ref{sub:reward}) and the number of consecutive grasping failures $c_g$. This knowledge provides direct information for the classifier to make action decisions: 1) $r_b$ and $n_b$ are indicators of the clutteredness around the target but hard for the networks to learn directly and 2) $c_g$ records the history of failed grasps.

We train the classifier to determine whether to push or grasp under the given state. The action with maximum corresponding Q value is executed. We name the classifier as action classifier ${f_a}$ (i.e., classify the state into grasp-favored or not) and the coordinator is formulated as
\begin{align}
    y &= \underset{\{p,g\}}{f_a}\!(q_p,q_g,r_b,n_b,c_g)\\
    \pi_c &: \argmax_a Q_y
\end{align}
where $f_a$ is a function approximator composed of three fully connected layers with batch normalization \cite{ioffe2015batch} and ReLU \cite{nair2010rectified}. It learns to signify the influential variables through its weights and drop unimportant factors by ReLU.

\begin{algorithm}
\caption{Critic-Policy to Grasping the Invisible}
\textbf{Input}: RGB-D image $I$\\
\textbf{Output}: action decision $a_t$ at time $t$
\begin{algorithmic}[1]
\State $M$ $\gets$ \texttt{ObjectSegmentation}($I$)
\If{$M=$\O}
\Comment{exploration subtask}
\State $M$ $\gets$ \texttt{AllOnesMask}()
\State $s_t$ $\gets$ \texttt{HeightmapProjection}($I$,$M$)
\State $C_p \gets$ \texttt{ClutterPrior}($s_t$)
\State $A_p \gets \phi_p(s_t)$
\State $F_p \gets$ \texttt{FailureLikelihood}$(\cdot)$
\State $a_t$ $\gets \underset{a}\argmax \, F_p \circ (C_p \circ A_p)$
\Comment{explorer}
\Else
\Comment{coordination subtask}
\State $s_t$ $\gets$ \texttt{HeightmapProjection}($I$,$M$)
\State $Q_p \gets \phi_p(s_t)$, $Q_g \gets \phi_g(s_t)$
\State $y \gets \! \underset{\{p,g\}}{f_a}\!(\max{Q_p},\max{Q_g},r_b,n_b,c_g)$
\State $a_t \gets \underset{a}\argmax \, Q_y$
\Comment{coordinator}
\EndIf
\end{algorithmic}
\label{alg:testing}
\end{algorithm}

\subsection{Policy Execution}
Algorithm \ref{alg:testing} summarizes the details of policy execution to grasping the invisible target. The algorithm is repeated until the robot grasps the target or exceeds the maximum number of motions. For each iteration, one of the two policies (explorer or coordinator) is effective upon the existence of target mask $M$.

\section{Learning Target-oriented Motion critic}
With the setup of visible targets, our approach learns the target-oriented motion critic, which is used in both explorer $\pi_e$ and coordinator $\pi_c$. In what follows, we discuss the target-oriented motion critic and the reward scheme for training.

\subsection{Motion critic}\label{sec:critic}
As discussed in Sec. \ref{sec:introduction}, when the visible target is within clutter, a sequence of target-oriented pushing and grasping actions should be applied to free the space around the target, and finally grasp it. With the RGB-D-mask heightmaps of the scene, it is sufficient to estimate the state of the current workspace. More specifically, a reasonable estimate of the target pose can be derived. Thus, we model the target-oriented manipulation problem as a discrete Markov Decision Process (MDP). For convenience of notations, we treat the heightmaps equivalent to the state, i.e., $s_t = (c_t,d_t,m_t)$.

In MDP, the robot performs an action $a_t$ in state $s_t$, then transitions to state $s_{t+1}$, and receives the corresponding reward $R(s_t,a_t,s_{t+1})$. The goal of our critic is to learn the action-value function $Q_{\{p,g\}}^\pi(s,a)$ which predicts the expected return for pushing or grasping action $a$ in state $s$ under a policy $\pi$. 

The motion critic is represented with a fully convolutional encoder-decoder network \cite{badrinarayanan2017segnet}, and it outputs pixel-wise Q maps. The RGB, depth, and mask heightmaps are fed into the corresponding feature extractor (2-layer residual \cite{he2016deep} network) for feature extraction. The extracted features are concatenated as the input to a DenseNet-121 \cite{huang2017densely} pre-trained on ImageNet \cite{deng2009imagenet}, to produce the motion-agnostic features \cite{pinto2017learning}. Then push network $\phi_p$ and grasp network $\phi_g$ take the features as input to predict push maps $Q_p$ and grasp maps $Q_g$, respectively. Network $\phi_p$ and $\phi_g$ have the same architecture, a 3-layer residual network followed by bilinear upsampling.

\subsection{Reward Function} \label{sub:reward}
We divide our reward scheme for the critic into two stages: the pre-action stage and post-action stage. For one action, either zero reward or maximum stage reward is assigned.

\begin{figure}[t]
    \centering
    \begin{subfigure}{0.11\textwidth}
        \centering
        \includegraphics[width=\textwidth]{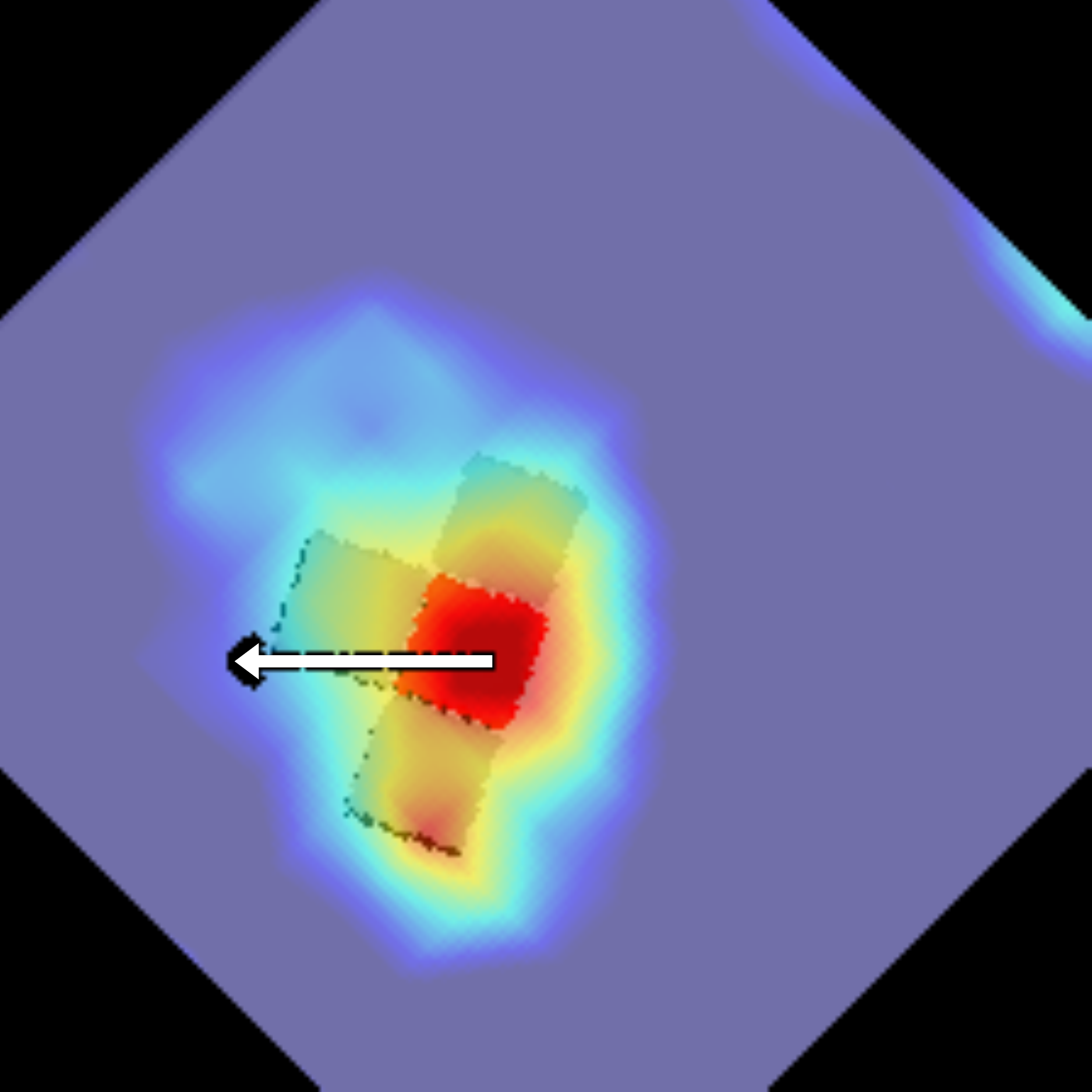}
        \caption*{$R_p=0.25$}
    \end{subfigure}
    \begin{subfigure}{0.11\textwidth}
        \centering
        \includegraphics[width=\textwidth]{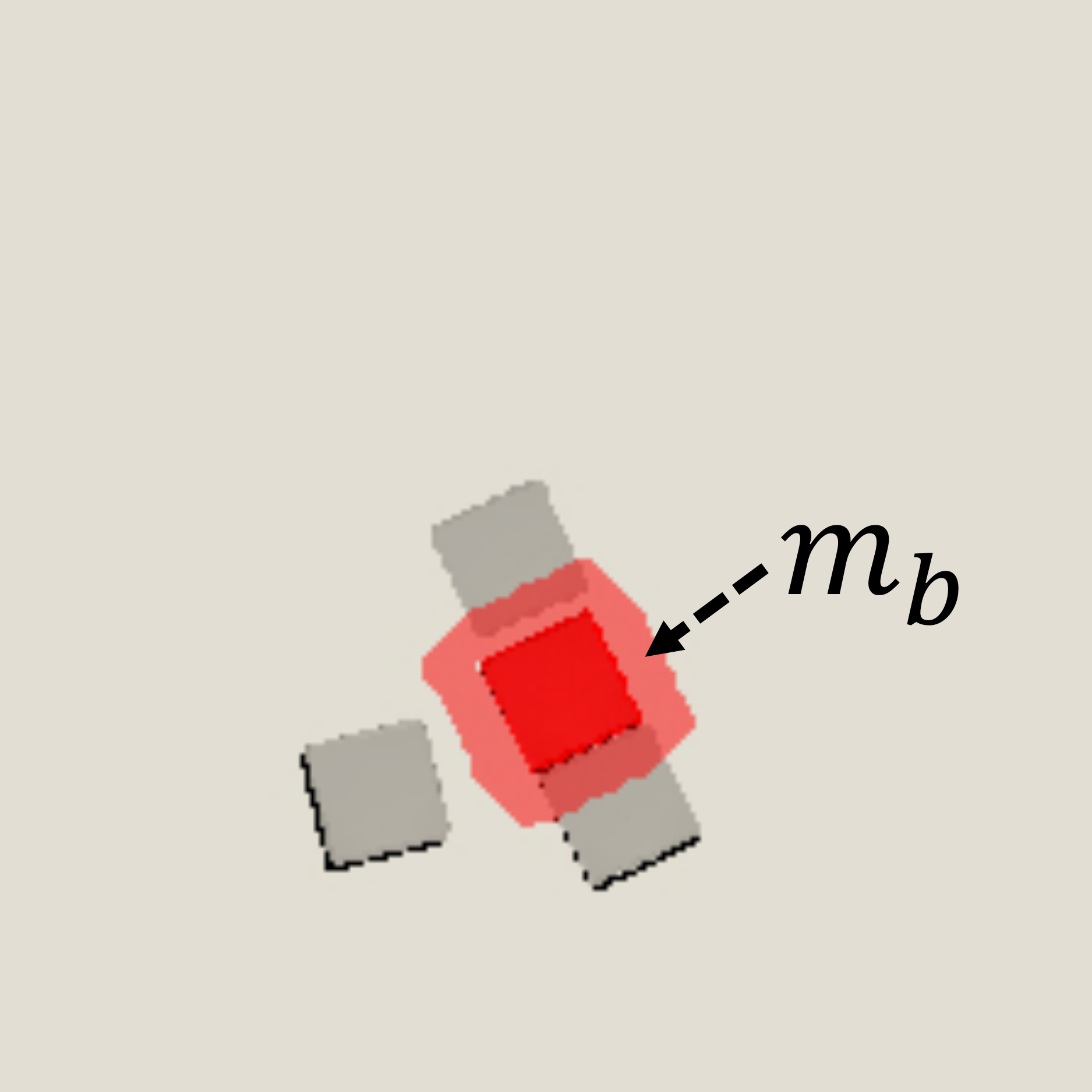}
        \caption*{$R_p=0.5$}
    \end{subfigure}
    \begin{subfigure}{0.11\textwidth}
        \centering
        \includegraphics[width=\textwidth]{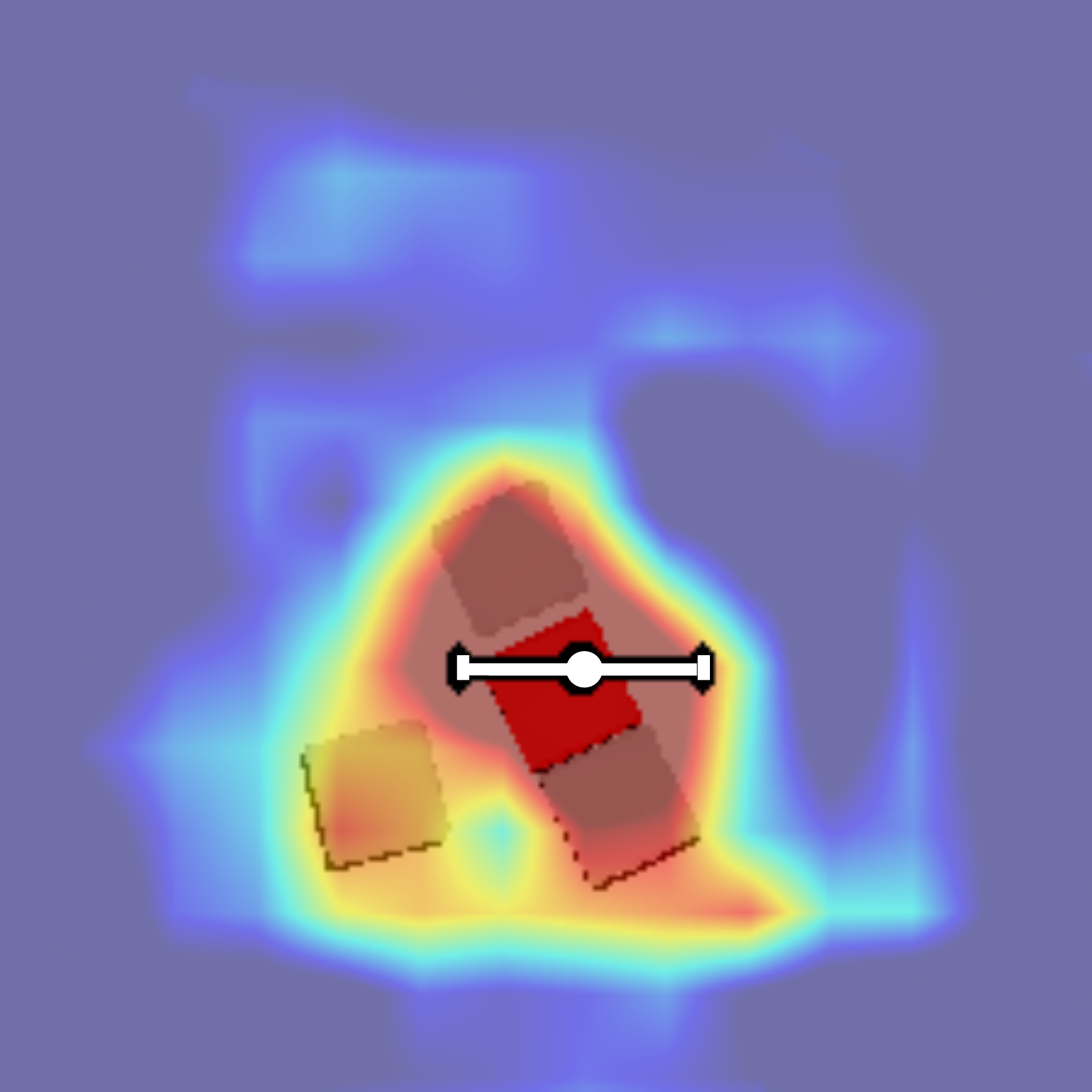}
        \caption*{$R_g=0.5$}
    \end{subfigure}
    \begin{subfigure}{0.11\textwidth}
        \centering
        \includegraphics[width=\textwidth]{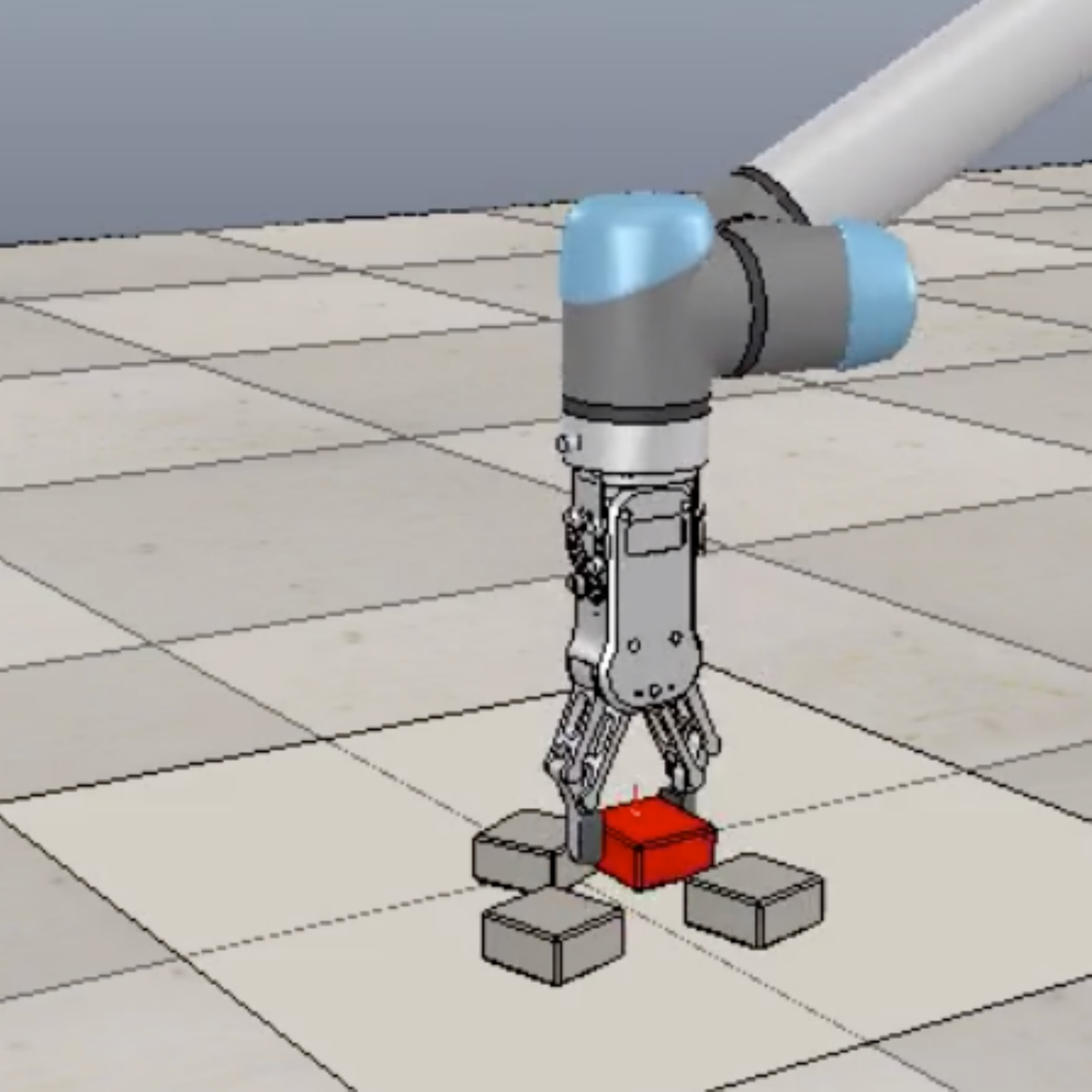}
        \caption*{$R_g=1$}
    \end{subfigure}
    \vspace{-2pt}
    \caption{\textbf{Example of our reward scheme}. One push and grasp are executed consecutively for the target (the red cuboid), and we assign the pre-action and post-action rewards to the corresponding actions.}
    \vspace{-5pt}
    \label{fig:rewards}
\end{figure}

We give an example of the reward scheme in Fig. \ref{fig:rewards}. We assign pre-action reward $R_p(s_t,s_{t+1}) = 0.25$ if the intended pushing vector passes mask $m_t$. Then post-action reward $R_p(s_t,s_{t+1}) = 0.5$ is given for pushes that make more space around the target for future grasping. To detect the space increase, we first dilate around $m_t$ to construct the mask of target border $m_b$ (shown as the mask of light red color). And the space increase is confirmed if border occupancy value $o_b$ (defined as the number of pixels in $m_b$ with height above the ground) decreased by some threshold. In the example, the pushing action frees spaces around the target, and thus the reward of 0.5 is given.

Similarly, we assign pre-action reward $R_g(s_t,s_{t+1}) = 0.5$ for those grasps with an intended grasping position in $m_t$, and post-action reward $R_g(s_t,s_{t+1}) = 1$ if the target is successfully grasped.

\section{Training in Self-supervision}
We train the motion critic and action classifier $f_a$ of policy $\pi_c$ by self-supervision in simulation.
\subsection{Loss Function}
The motion critic is trained by minimizing temporal difference error $\delta_t$ as
\begin{align}
    &\delta_t = Q(\theta_t; s_t,a_t)-(R_{a_t}(s_t,s_{t+1})+\gamma \max_a Q(\theta_t^{-}; s_{t+1},a))
\end{align}
via the Huber loss
\begin{align}
    &\mathcal{L}_{\delta}=
    \begin{cases}
      \frac{1}{2} \delta_t^2, & \text{if}\ |\delta_t| \leq 1 \\
      |\delta_t|-\frac{1}{2}, & \text{otherwise}
    \end{cases}
\end{align}
where $\theta_t$ are the parameters of the critic networks at time $t$, and the target network parameters $\theta_t^{-}$ are held fixed between iterations. At time $t$, we pass gradients only through the single pixel on which the motion primitive was executed while all other pixels backpropagate with 0 loss.

The coordinator is trained using the binary cross-entropy loss
\begin{align}
    \label{eq:fa}
    \mathcal{L}_{y}=-(\bar{y} \log y+(1-\bar{y}) \log(1-y))
\end{align}
where $y$ is the predication from action classifier $f_a$ and $\bar{y}$ is the ground-truth label.

\subsection{Data Collection and Training}
We collect the data with the following procedure: $n$ target candidates (i.e., detectable by the semantic segmentation module) and $m$ basic objects are randomly selected and dropped into the workspace in front of the robot. The robot needs to grasp one randomly appointed target via a sequence of pushing and grasping. Once the target is successfully grasped, the new target is appointed for the next trial. The objects are again randomly dropped if the workspace is void of target candidates. We save heightmaps, executed actions, and execution results for training the critic. In addition, labels are automatically generated for grasping actions to train policy $\pi_c$ (see equation \ref{eq:fa}). The label $\bar{y}$ is assigned as 1 if the target is grasped or 0 if the grasping position is within mask $m_t$ but results in a grasping failure (this might indicate dense clutter around the target). 

Multi-stage training is utilized. At the first stage, we only train the critic to reach a good initialization; the robot follows under an $\epsilon$-greedy policy $\pi_\epsilon$. We set $m=3$ to ease the learning of target-oriented pushing and grasping. Then $m$ increases to be 8, and the policy switches to be randomly initialized coordinator $\pi_c$ to learn coordinated decision making in structured dense clutter. In the meantime, the critic is still under training and expected to accommodate $\pi_c$, i.e., fine-tuned from $Q^{\pi_\epsilon}$ to $Q^{\pi_c}$. We summarize the training process in Appendix \ref{appen:train}.
\begin{figure}[t]
  \centering
  \includegraphics[width=0.48\textwidth]{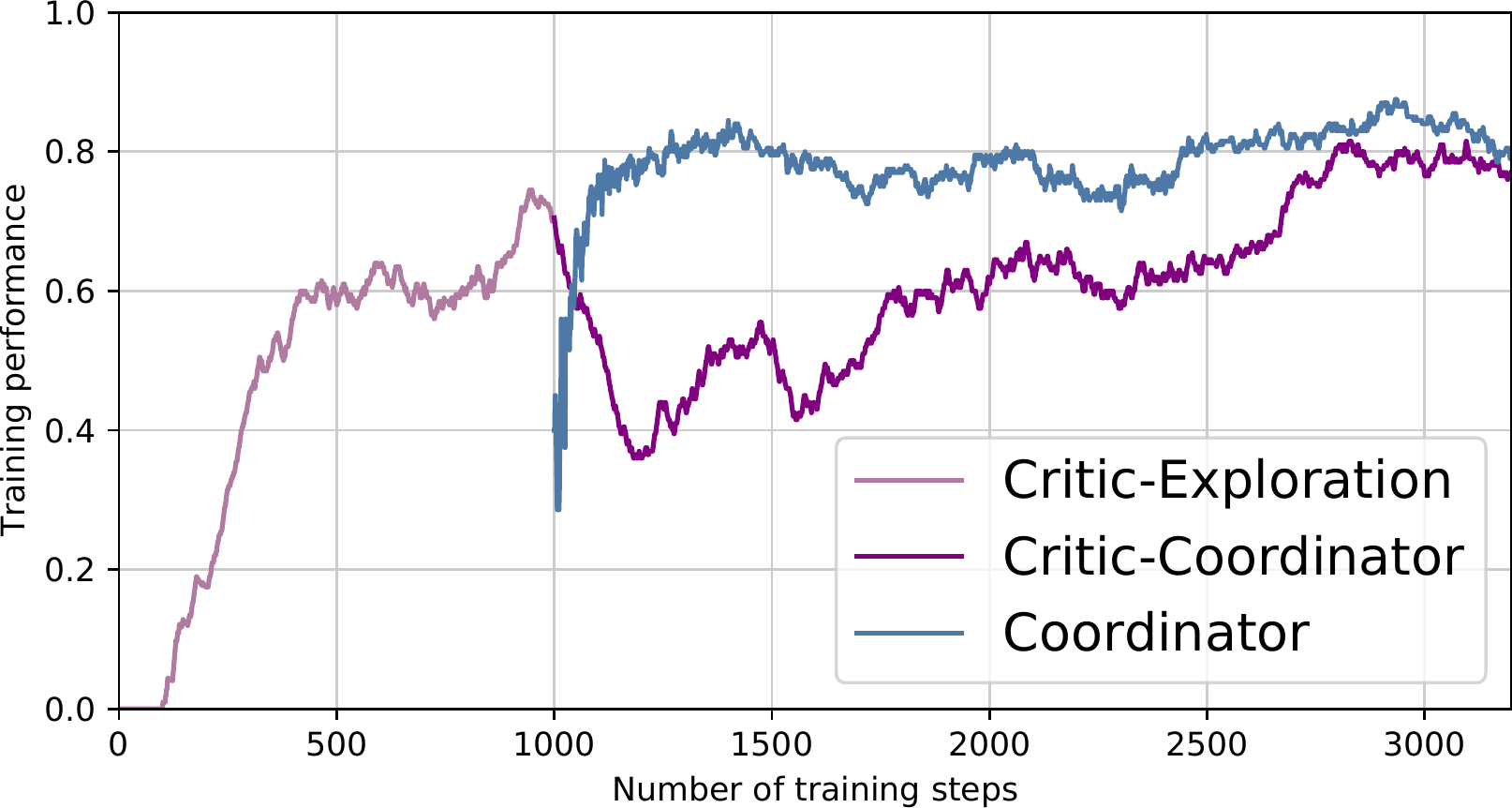}
  \vspace{-3pt}
  \caption{\textbf{Training performance.} The purple lines indicate the target-oriented grasping success rate of the critic-policy, and the blue line indicates the training accuracy of action classifier $f_a$ in policy $\pi_c$ over training steps.}
  \label{fig:training}
  \vspace{-3pt}
\end{figure}

As shown in Fig. \ref{fig:training}, only the critic is trained at the first stage (first 1000 iterations in our experiments) under $\pi_\epsilon$ exploration and reaches a high target-oriented grasping success rate (defined as $\frac{\#\text{ successful target-oriented grasping}}{\#\text{ total motions (pushes and grasps)}}$). Then we replace the policy to be coordinator $\pi_c$ and start to train the coordinator to increase its prediction accuracy gradually. Note that critic-coordinator finally achieves a higher target-oriented grasping success rate in even more cluttered scenes.

\begin{figure*}[t]
  \centering
    \begin{subfigure}{0.39\textwidth}
        \centering
        \includegraphics[scale=0.4]{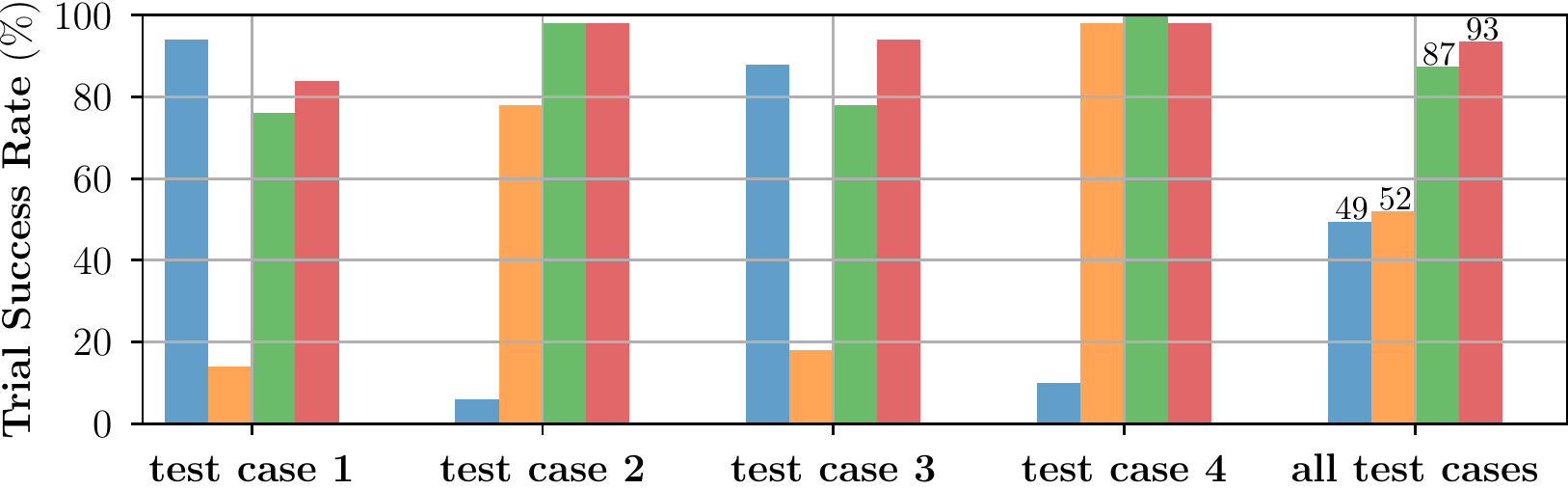}
    \end{subfigure}
    \hfill
    \begin{subfigure}{0.6\textwidth}
        \centering
        \includegraphics[scale=0.4]{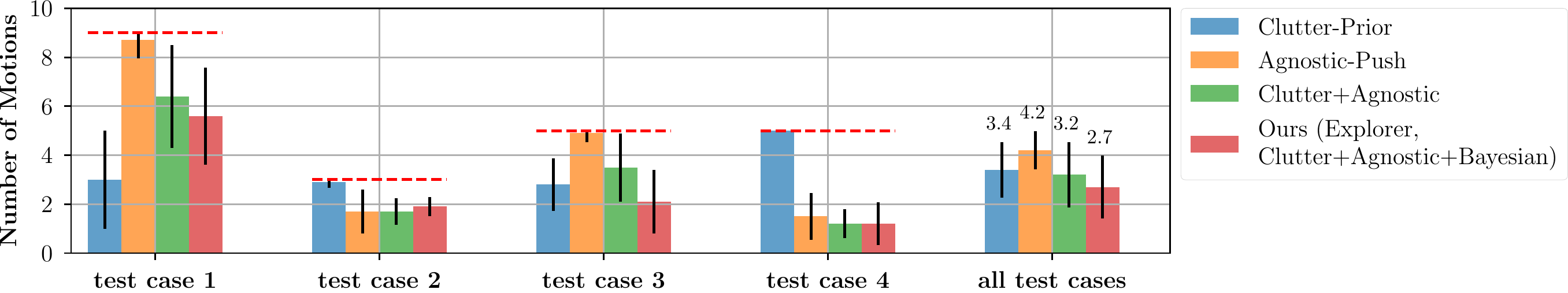}
    \end{subfigure}
  \caption{\textbf{Performance in exploration subtask.} The task success rate (left) and the number of motions (right) of four approaches on the four test cases of complete occlusion. The plot shows the effectiveness of our approach, \textbf{Explorer}, which achieves a task success rate of $93\%$ with 2.7 motions on average. The red dotted line is the number limitation of motions.}
  \vspace{-5pt}
  \label{fig:bar_explore_trail}
\end{figure*}

\section{EXPERIMENTS}
We train the system in simulation where all states are known. We executed a set of ablation studies for explorer $\pi_e$ and comparative experiments for coordinator $\pi_c$. The goals of the experiments are 1) to show the importance of the extra domain knowledge in the policies, 2) to understand the limitation of each submodule in explorer $\pi_e$ and the advantages and robustness of $\pi_e$ and 3) to demonstrate that our coordinator $\pi_c$ can coordinate target-oriented pushing and grasping in structured clutter. The simulation environment and the robot are kept the same with \cite{zeng2018learning} for a fair comparison. We also run experiments on a real robot to show the performance of our system on the ``grasping the invisible'' problem. The testing objects are either toy blocks or daily objects of a similar shape to the training objects. Although it would be of interest to see how our system generalizes to quite different testing objects, the main goal of the experiments in this paper is to show the generalization of our approach to challenging object arrangements never seen during training.

\subsection{Exploration Subtask} \label{sub:experiments_exploration}
\begin{figure}[t]
    \centering
    \begin{subfigure}{0.48\textwidth}
        {\includegraphics[width=0.24\textwidth]{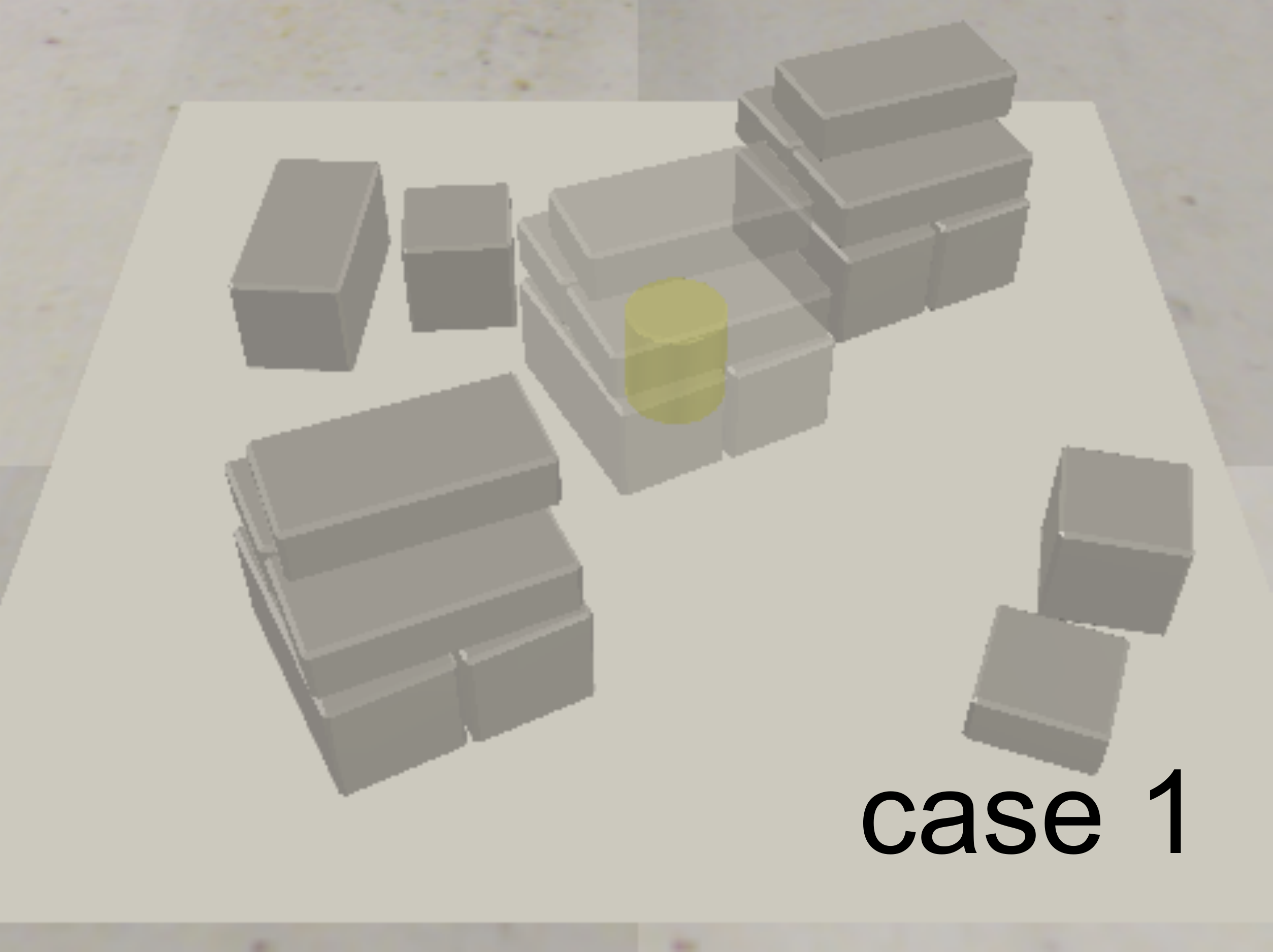}}\hfill
        {\includegraphics[width=0.24\textwidth]{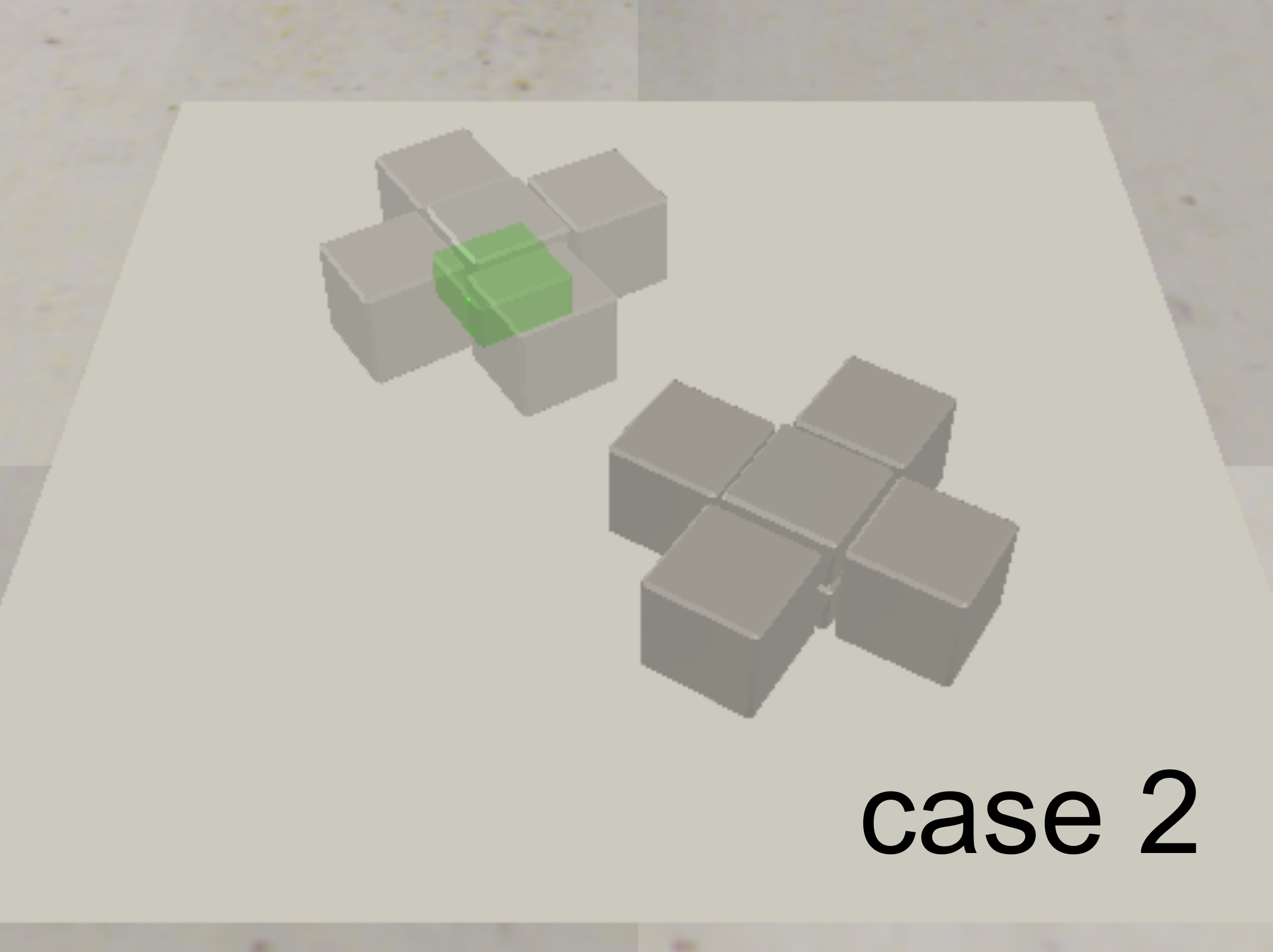}}\hfill
        {\includegraphics[width=0.24\textwidth]{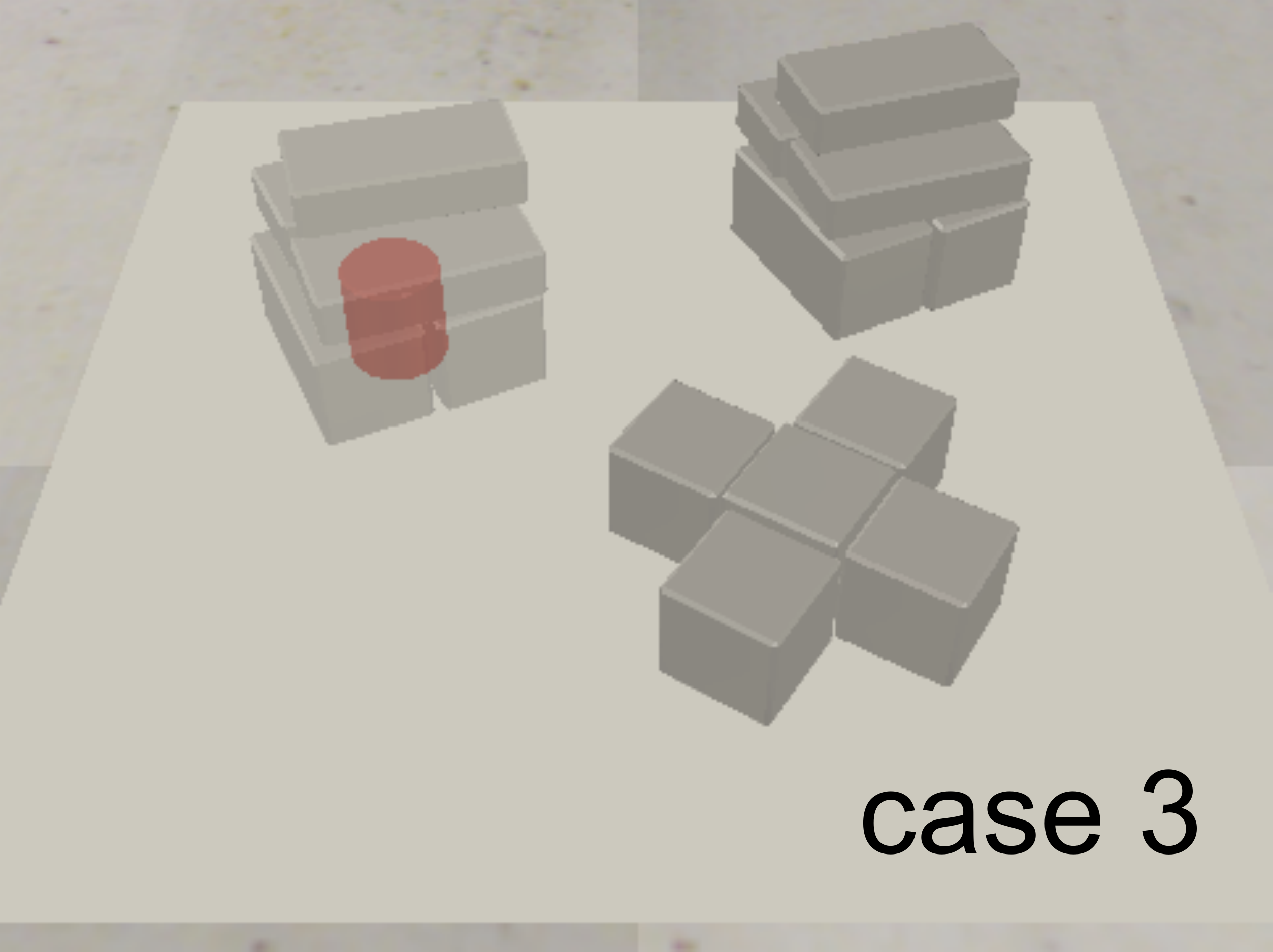}}\hfill
        {\includegraphics[width=0.24\textwidth]{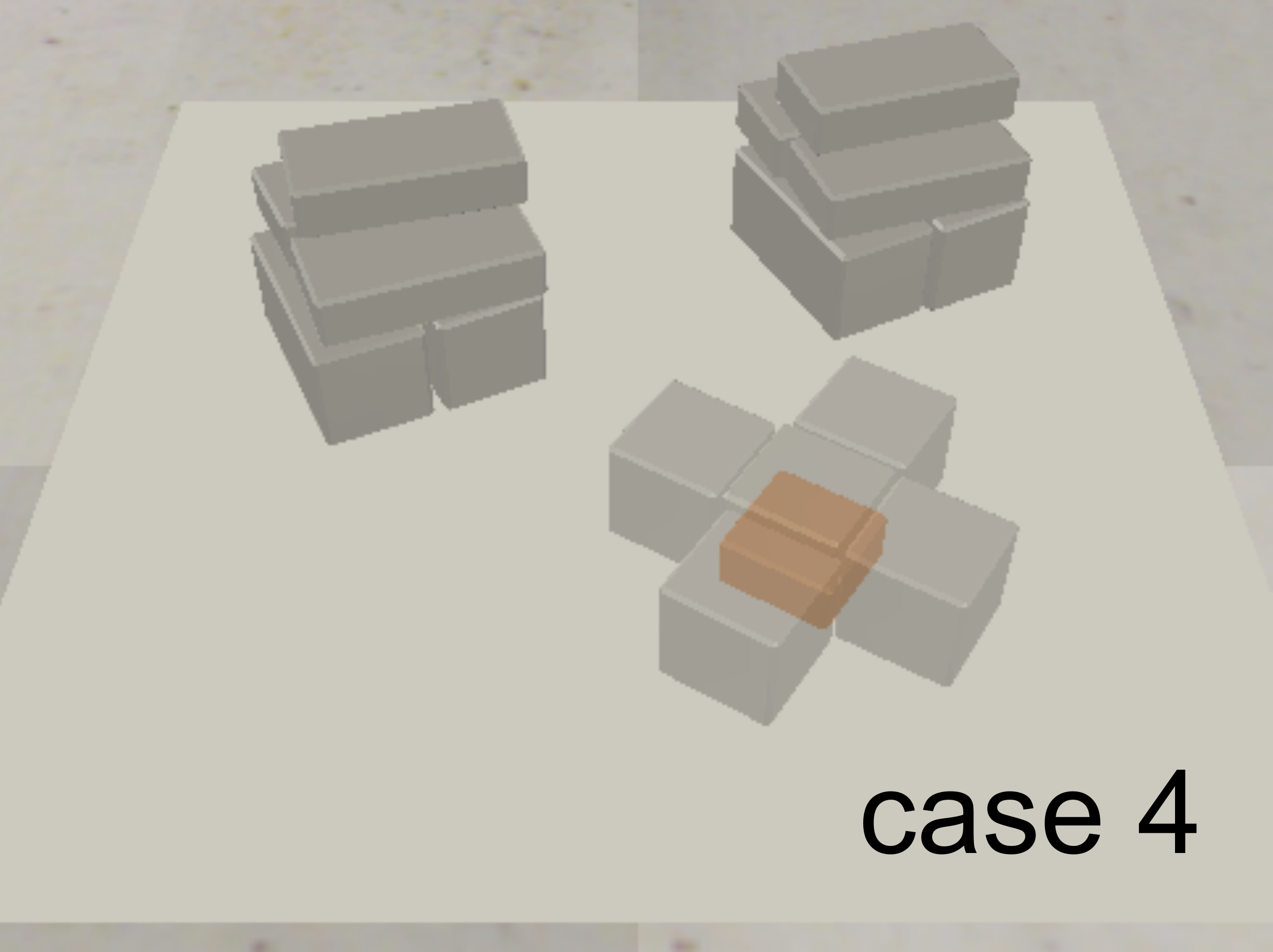}}%
        \vspace{-4pt}
        \caption{Test cases in exploration subtask}
        \vspace{2pt}
        \label{fig:invisible}
    \end{subfigure}
    \begin{subfigure}{0.48\textwidth}
        {\includegraphics[width=0.24\textwidth]{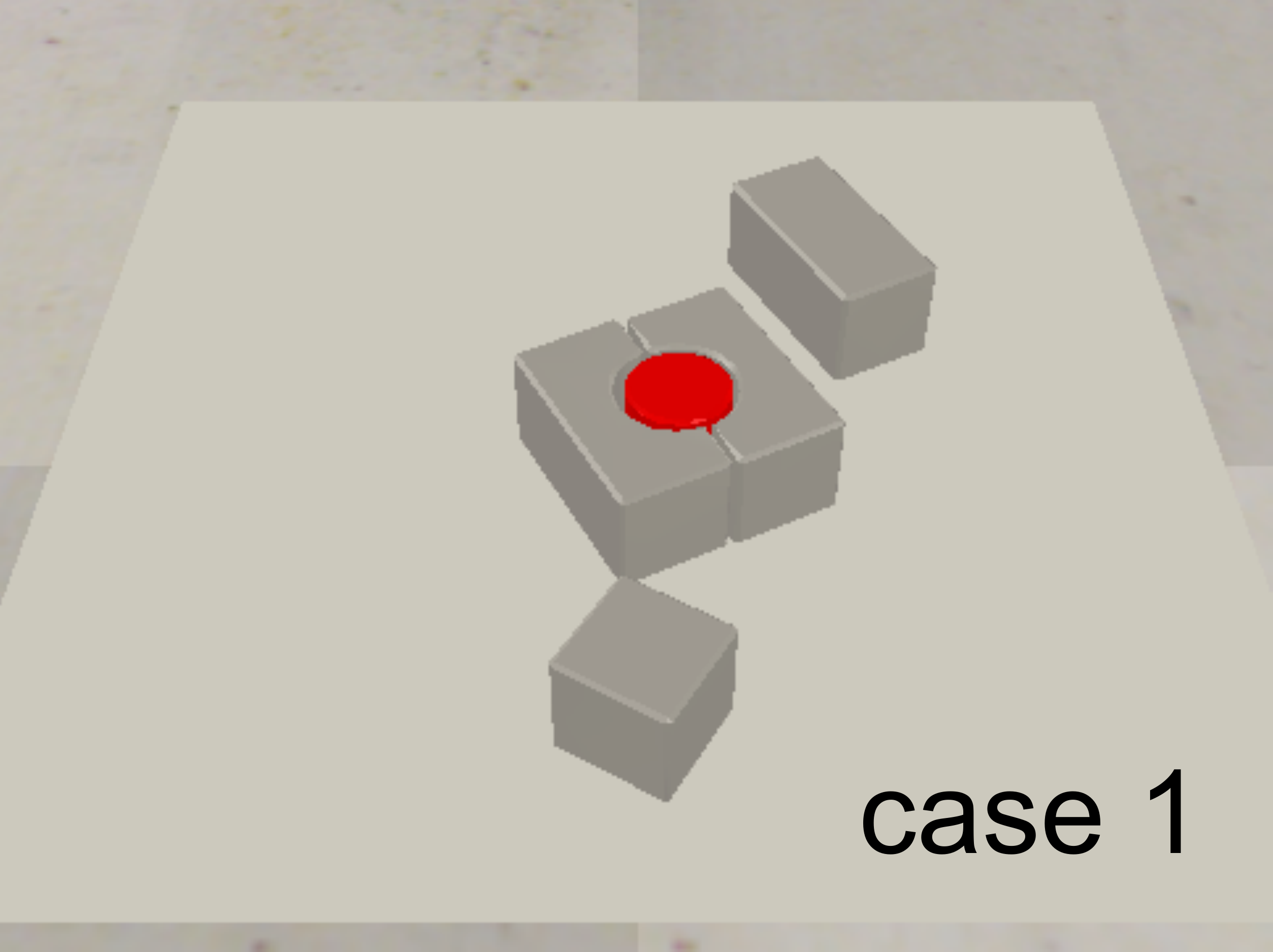}}\hfill
        {\includegraphics[width=0.24\textwidth]{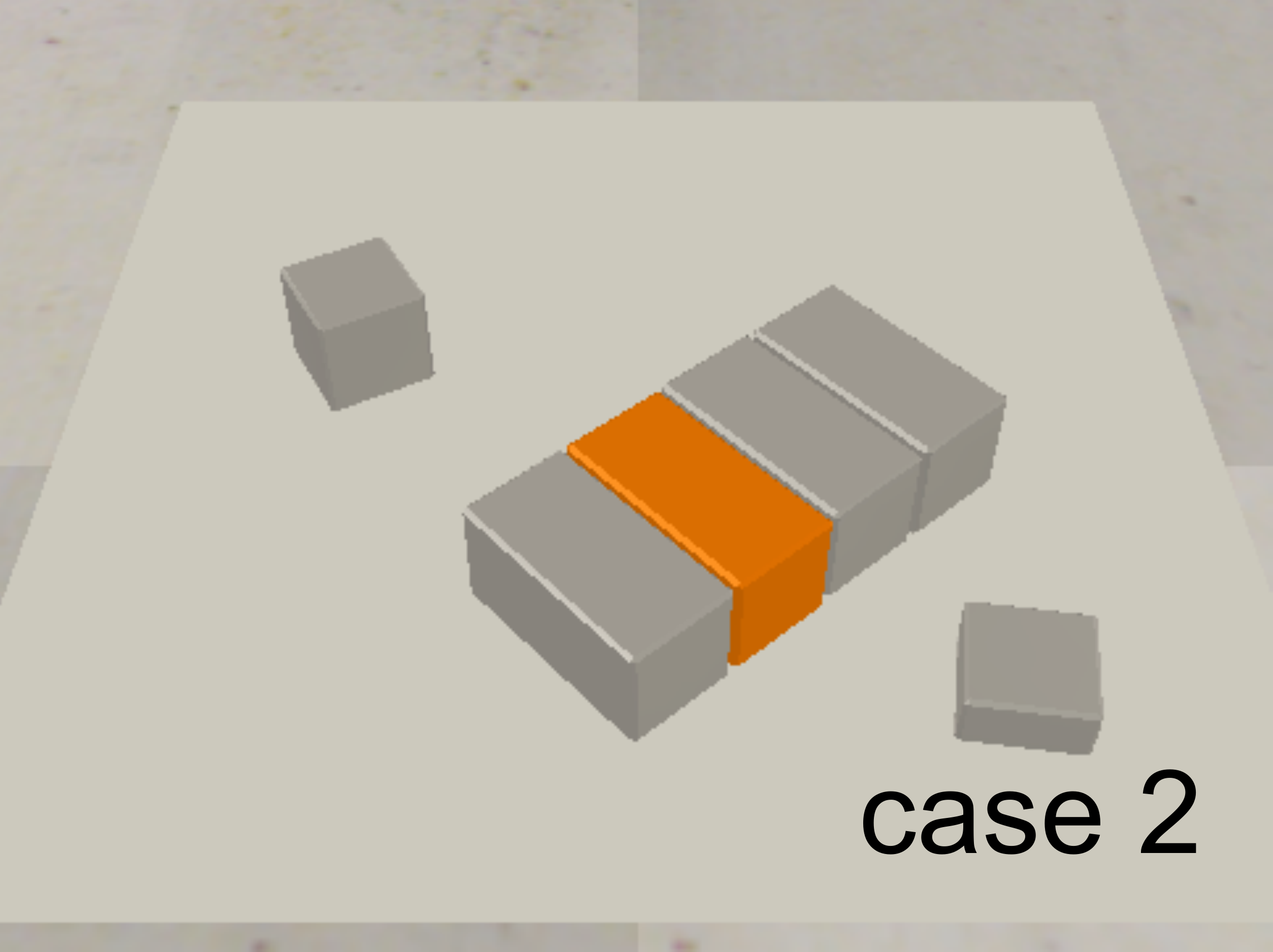}}\hfill
        {\includegraphics[width=0.24\textwidth]{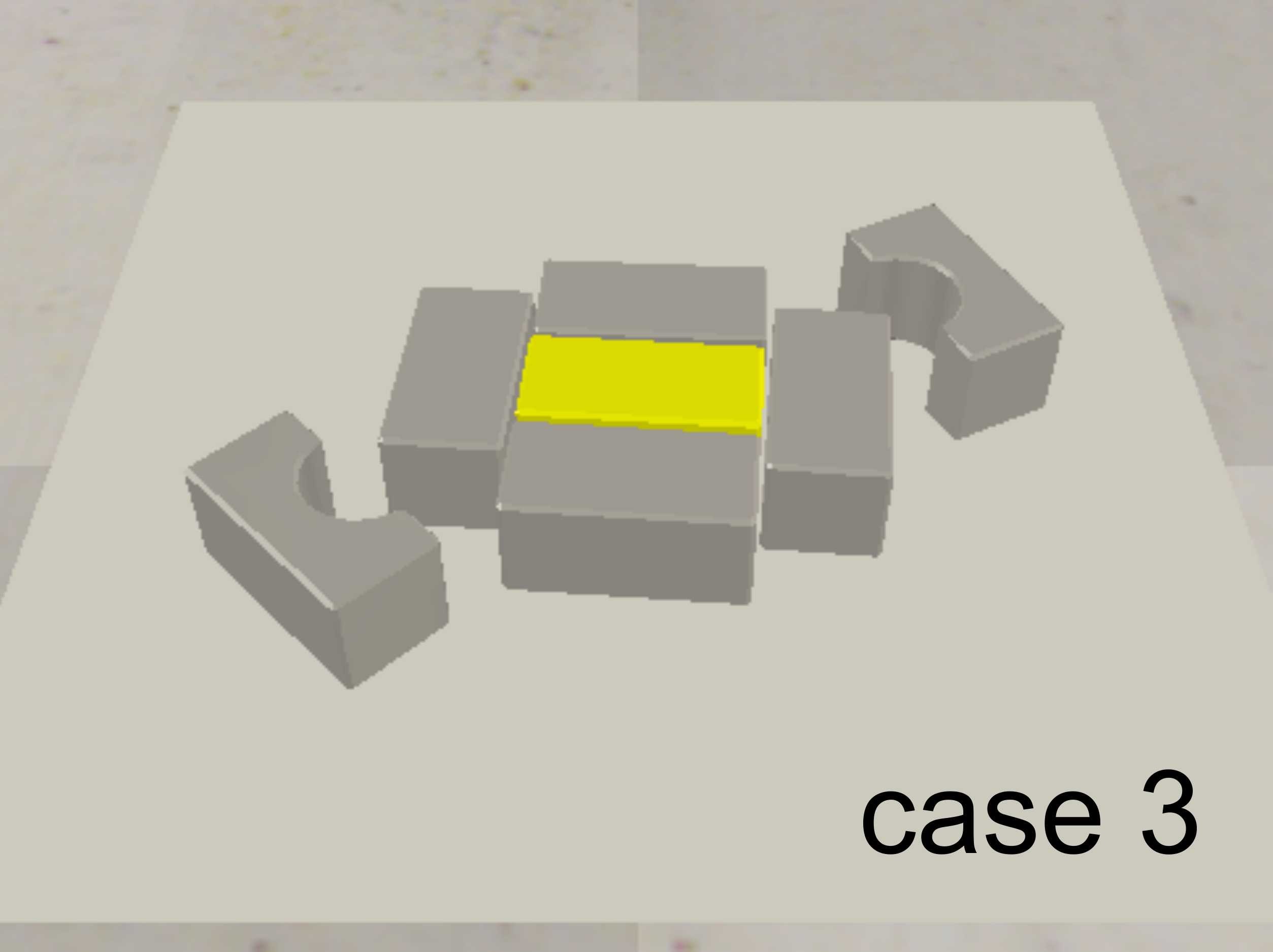}}\hfill
        {\includegraphics[width=0.24\textwidth]{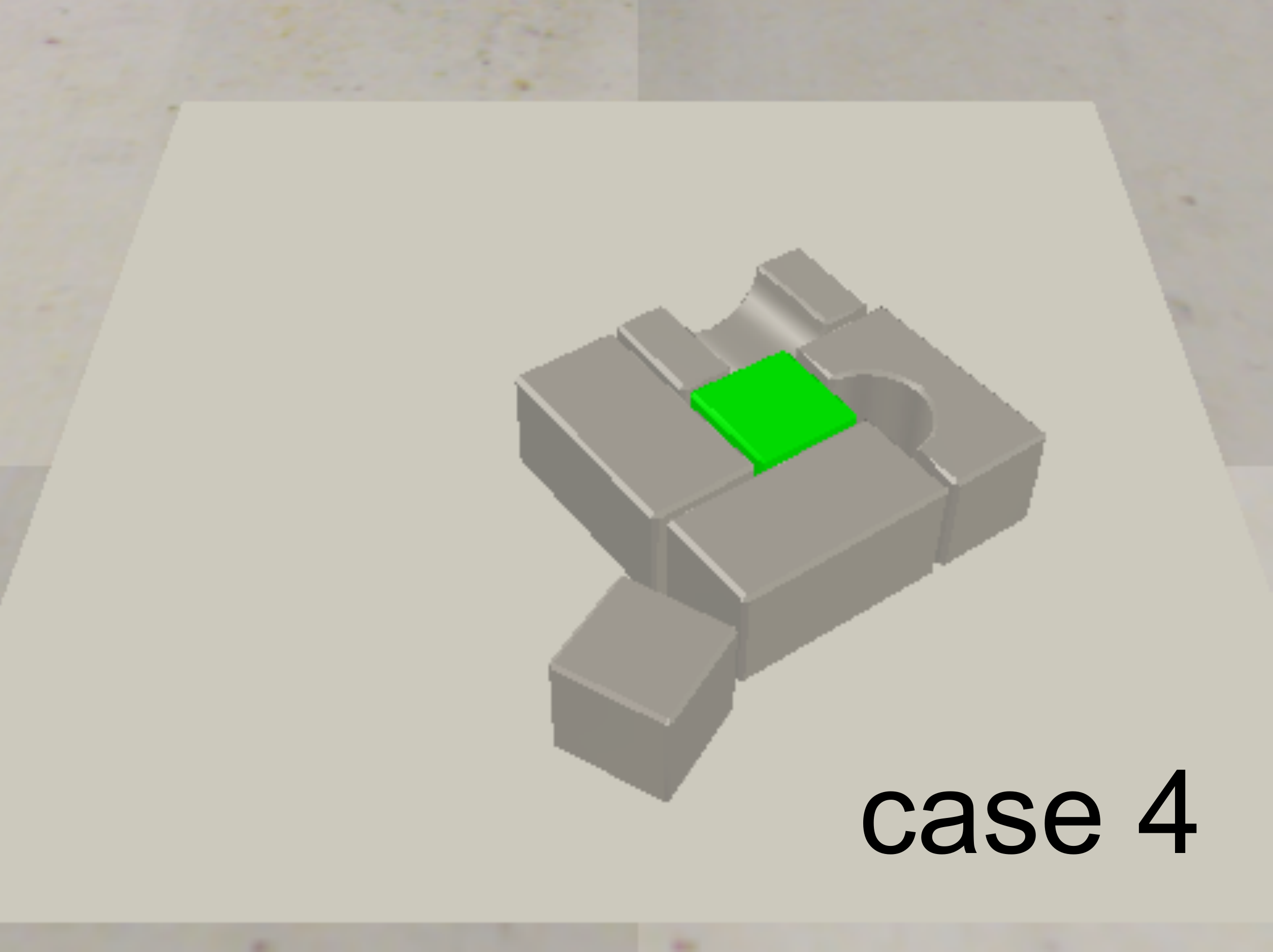}}%
        \\[\smallskipamount]
        {\includegraphics[width=0.24\textwidth]{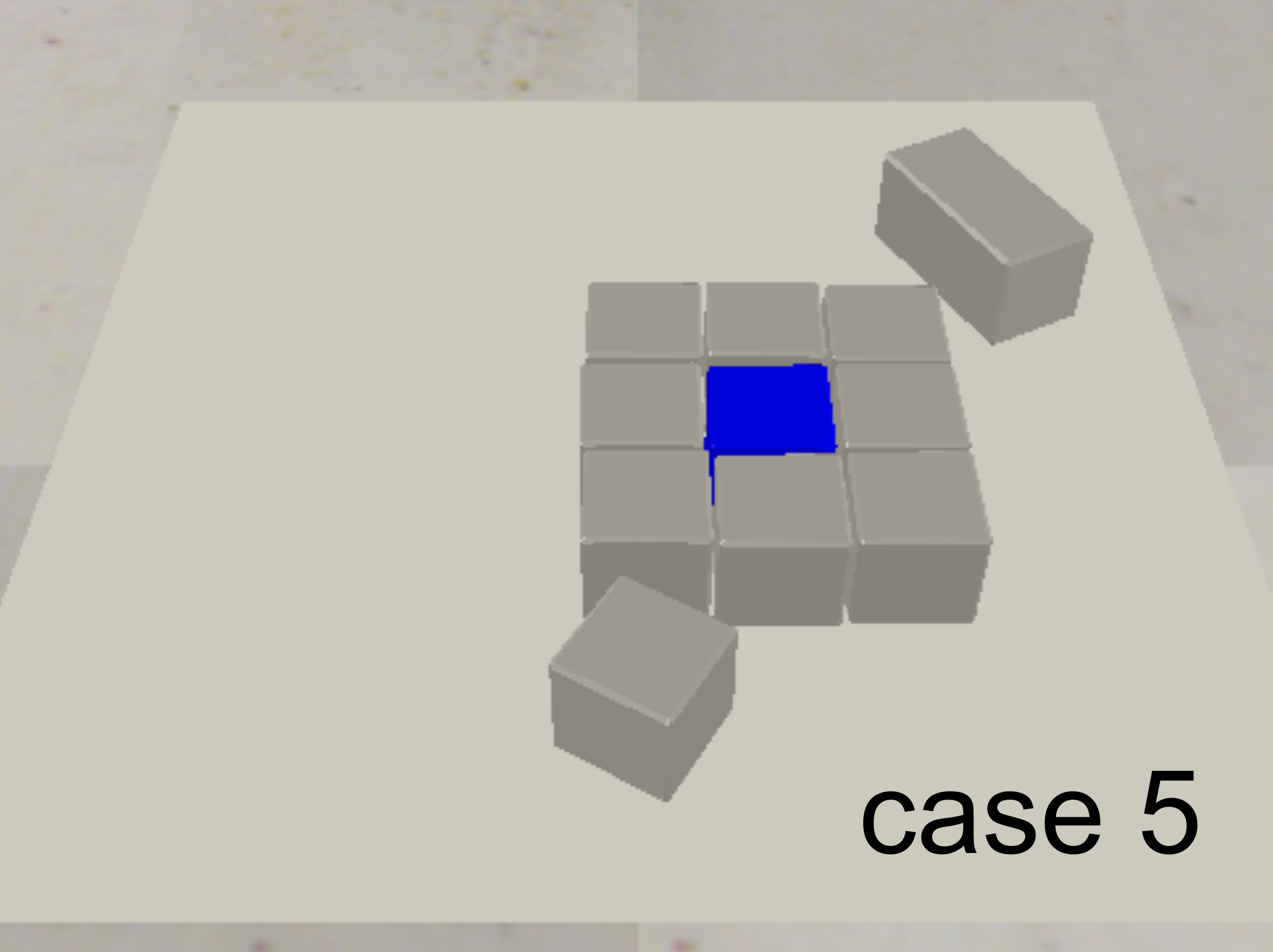}}\hfill
        {\includegraphics[width=0.24\textwidth]{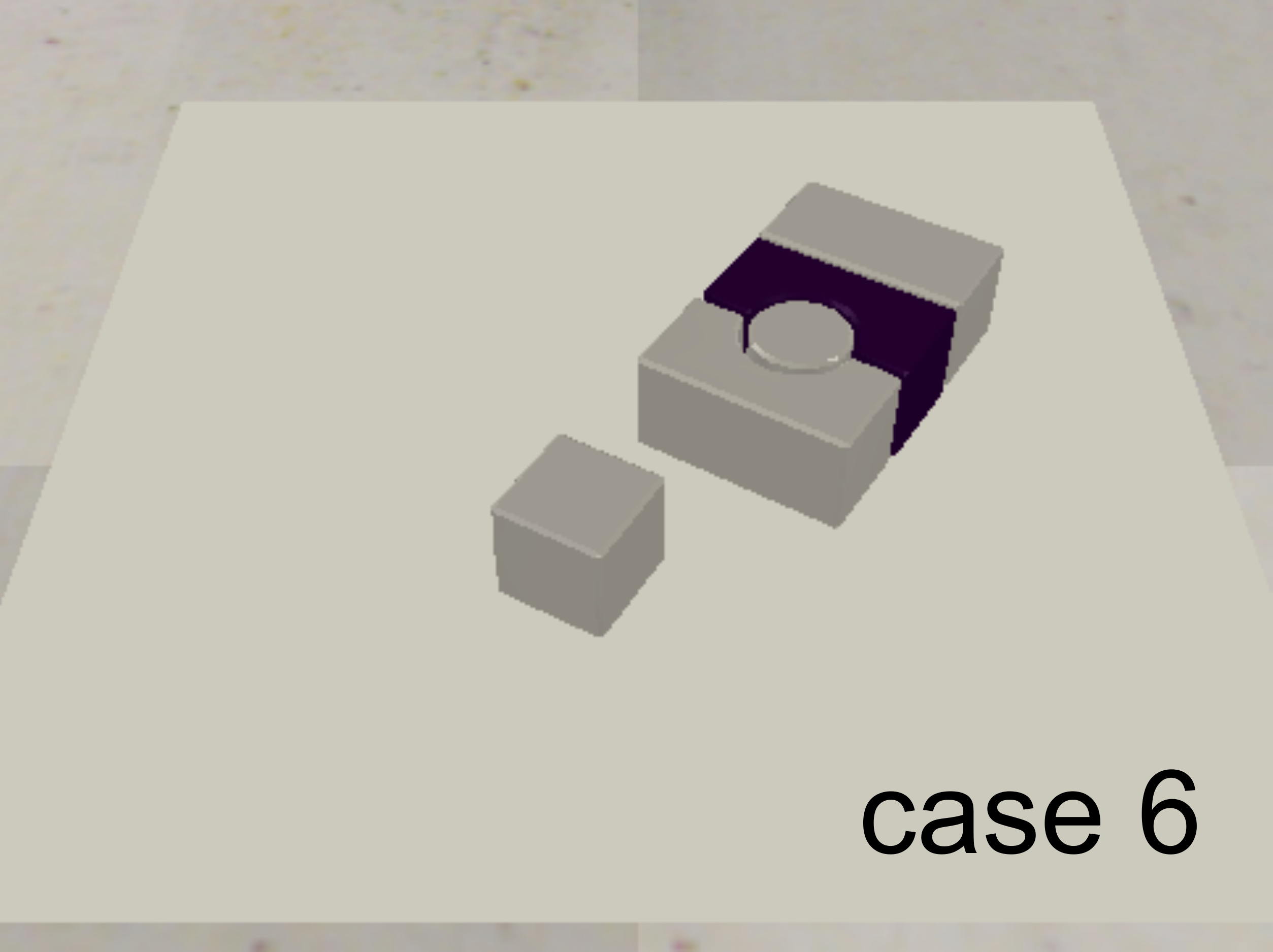}}\hfill
        {\includegraphics[width=0.24\textwidth]{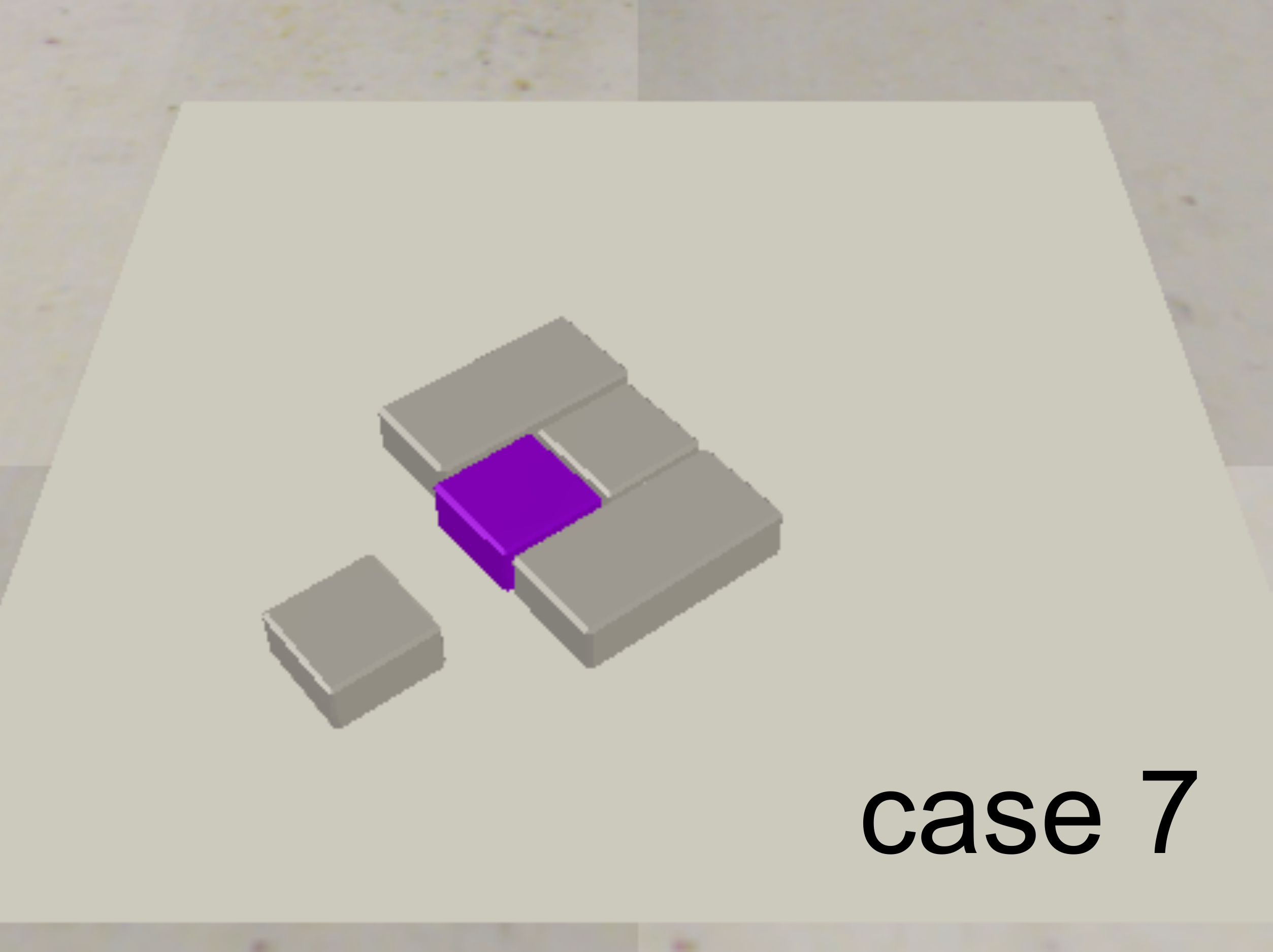}}\hfill
        {\includegraphics[width=0.24\textwidth]{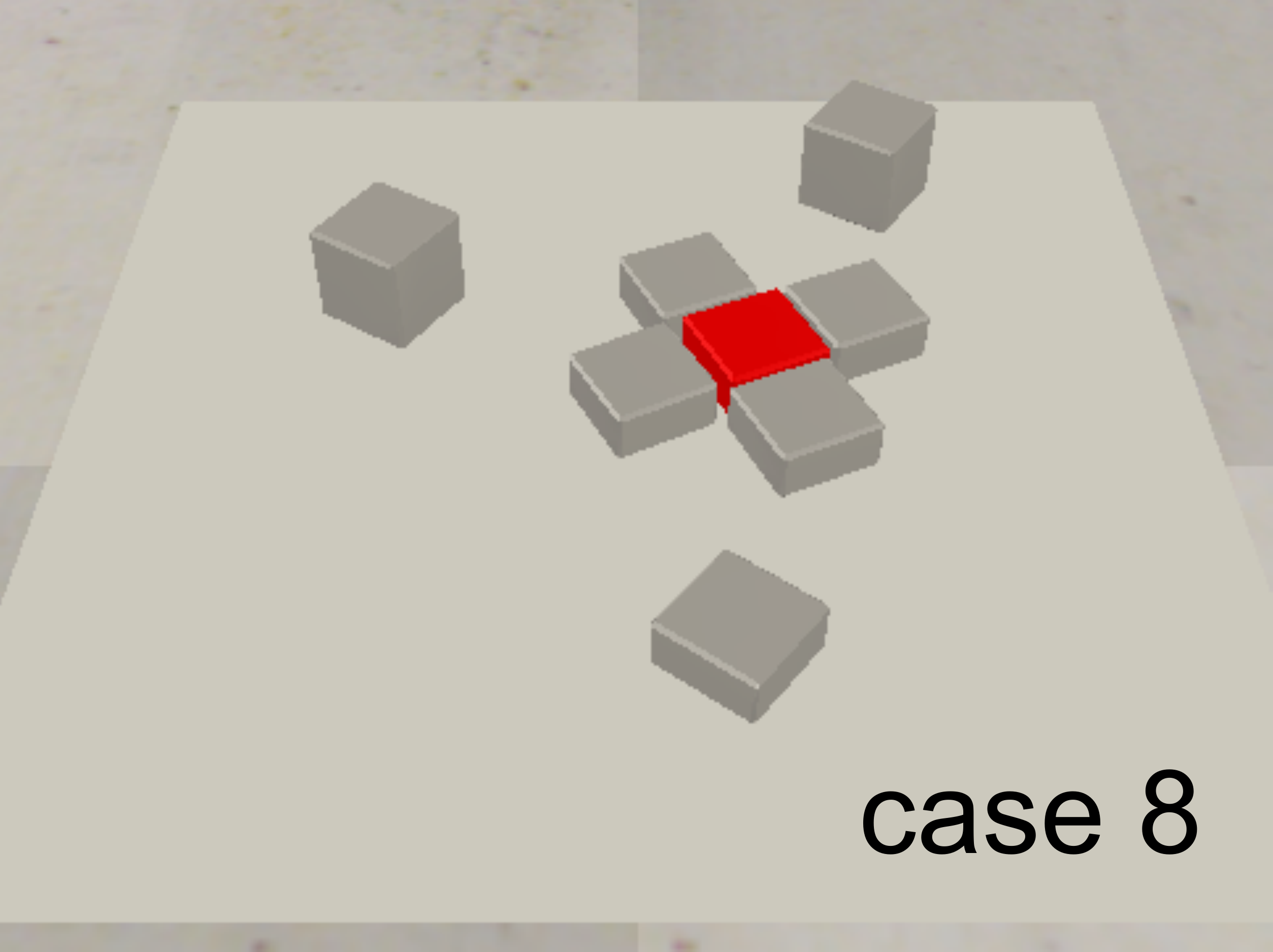}}%
        \vspace{-4pt}
        \caption{Test cases in coordination subtask}
        \vspace{-3pt}
        \label{fig:structures}
    \end{subfigure}
    \caption{\textbf{Test cases in simulation.} (a) shows four test cases in the exploration subtask, where the target is invisible, and (b) shows eight challenging arrangements where the coordination between pushing and grasping is required. The target is the colored object.}%
    \vspace{-5pt}
\end{figure}

To validate our approach, we run an ablation study for which our explorer is compared with the following methods in the exploration subtask:
1) \textbf{Clutter-Prior} builds the probability maps $C_p$ for potential pushing actions based on the depth heightmap \cite{zeng2018vpg}. The heightmap is first translated along one fixed axis for 25 pixels (approximately twice the width of the closed gripper), then the pixel with enough depth difference between original and translated heightmap is recorded as 1 otherwise 0. This binary map is filtered with a 25$\times$25 all-ones kernel to get a pixel-wise probability map. Like Q maps, the heightmap is also rotated by $N$ orientations to construct $N$ probability maps. By detecting varying heights, $C_p$ encodes prior knowledge of edges of clutter.
2) \textbf{Agnostic-Push} utilizes target-agnostic pushing $A_p$ from $\phi_p$ taking in RGB-D heightmaps $c_t$, $d_t$ and all-ones mask heightmap. \textbf{Agnostic-Push} lacks the sense of clutter and could push a well-isolated object, as $\phi_p$ is trained to make changes around the visible target by pushing.
3) \textbf{Clutter+Agnostic} is the Hadamard product of \textbf{Clutter-Prior} and \textbf{Agnostic-Push}, i.e., $C_p \circ A_p$. By balancing between clutter prior and target-agnostic pushing, the robot achieves significant performance gain.
4) \textbf{Clutter+Agnostic+Bayesian} is our explorer. It adds pushing failure likelihood $F_p$ on \textbf{Clutter+Agnostic} to reduce the probability around the three most recent locations of failed pushes, which improves the robustness of the system by avoiding getting stuck.
\begin{figure*}[t]
    \begin{subfigure}{\textwidth}
        \hspace{1.3cm}
        \includegraphics[scale=0.4]{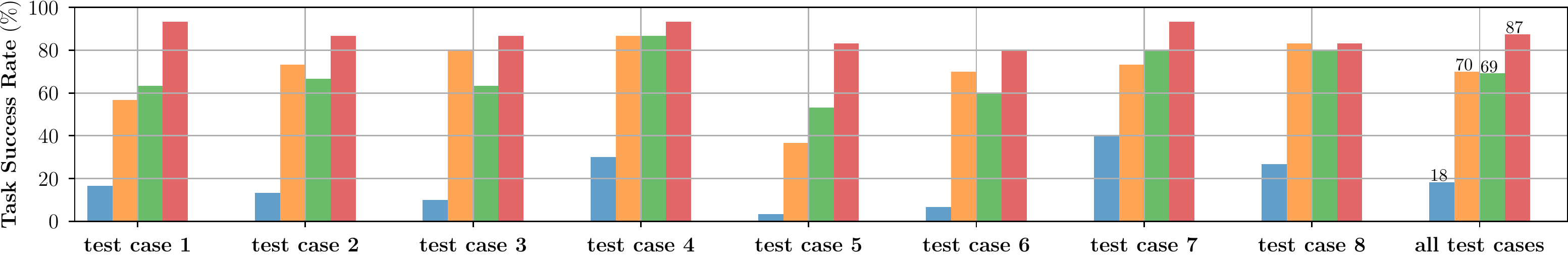}
    \end{subfigure}
    \vfill
    \begin{subfigure}{\textwidth}
        \hspace{1.32cm}
        \includegraphics[scale=0.4]{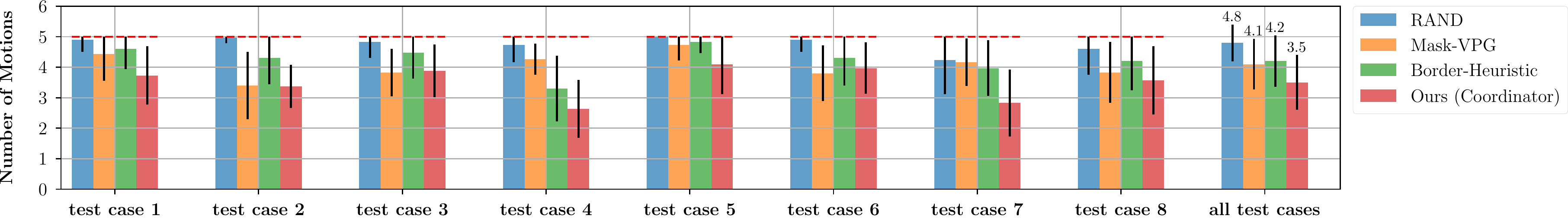}
    \end{subfigure}
  \caption{\textbf{Performance in coordination subtask.} The task success rate (top) and the number of motions (bottom) of four approaches on the 8 test cases of challenging arrangements. The plot shows the effectiveness of our approach, \textbf{Coordinator}, which achieves a task success rate of $87\%$ with 3.3 motions on average.}
  \vspace{-8pt}
  \label{fig:bar_coordinate_trail}
\end{figure*}
\begin{table}[t]
    \caption{Average Performance in the Exploration Subtask}
    \vspace{-2pt}
    \label{tab:average_exploration}
    \begin{tabular}{c|c|c}
    \hline
    Method & Task Success Rate ($\%$) & Number of Motions\\
    \hline
    Clutter-Prior & 49.5 & 3.43 $\pm$ 1.14\\
    Agnostic-Push & 52 & 4.20 $\pm$ 0.78\\
    Clutter+Agnostic & 87.5 & 3.20 $\pm$ 1.33\\
    Our Explorer & \textbf{93.5} & \textbf{2.70} $\pm$ 1.28\\
    \hline
    \end{tabular}
\end{table}

The test cases are shown in Fig. \ref{fig:invisible}. In each test case, the maximum number of pushes is $n_\text{pushes} = 2\ast n_\text{cluster}-1$ where $n_\text{cluster}$ is the number of object clusters in the workspace. The robot is tasked to find the target via explorational pushes and allowed to stop if the target is found. We execute 50 runs on each test case and report the task success rate and the number of motions. Fig. \ref{fig:bar_explore_trail} presents the performance of the above methods in the exploration subtask. Both \textbf{Clutter-Prior} and \textbf{Agnostic-Push} perform poorly in specific cases. For instance, \textbf{Clutter-Prior} is discouraged in either the equal height case (case 2) or the taller pyramid-like clutter case (case 4). \textbf{Agnostic-Push} is prone to push every object in the workspace until it finds the target buried under the pyramid-like shape in the test case 1 and 3, and hence it necessitates a large number of motions. By balancing between clutter prior and target-agnostic pushing, \textbf{Clutter+Agnostic} works consistently well across all test cases. Our explorer \textbf{Clutter+Agnostic+Bayesian} further improves the performance of \textbf{Clutter+Agnostic} in terms of the task success rate with a fewer number of motions. As shown in Table \ref{tab:average_exploration}, our explorer \textbf{Clutter+Agnostic+Bayesian} outperforms \textbf{Clutter+Agnostic} by $6\%$ in terms of the task success rate and requires about $0.5$ fewer average number of motions compared to \textbf{Clutter+Agnostic}.

\subsection{Coordination Subtask} \label{sub:experiments_coordination}

We compare the picking performance of our coordinator with the following baseline approaches:
1) \textbf{RAND} randomly chooses one of motion primitives and samples motion angle from the $N$ angles and motion position in target mask $m_t$.
2) \textbf{Mask-VPG} is an extension of VPG \cite{zeng2018learning} by incorporating the mask $m_t$ as post-processing. VPG produces Q maps for target-agnostic tasks, and we filter the push maps by a dilated target mask and the grasp maps by the target mask. The action with the highest Q value is executed.
3) \textbf{Border-Heuristic} determines whether to push or grasp by a heuristic policy and asks the motion critic to execute the action. The policy is $\epsilon$-greedy, and higher $\epsilon$ indicates a higher pushing exploration rate. The base rate is $\epsilon_0=0.5$, and we adjust its value to accommodate maximum Q values $q_p$, $q_g$ and the domain knowledge $r_b$, $n_b$, $c_g$ defined in Sec. \ref{sub:policy}. In brief $\epsilon$ is advanced with the decrease of $q_g-q_p$, the increase of $r_b$ and $n_b$, and the growth of $c_g$. Note that the method takes the same input with our coordinator, but the difference is the subtask policy. Please refer to Appendix \ref{appen:baseline} for more details.

\begin{table}[t]
    \caption{Average Performance in the Coordination Subtask}
    \vspace{-2pt}
    \label{tab:average_coordination}
    \begin{tabular}{c|c|c}
    \hline
    Method & Task Success Rate ($\%$) & Number of Motions\\
    \hline
    RAND & 18.3 & 4.77 $\pm$ 0.60\\
    Mask-VPG & 70.0 & 4.06 $\pm$ 0.83\\
    Border-Heuristic & 69.2 & 4.25 $\pm$ 0.85\\
    Our Coordinator & \textbf{87.5} & \textbf{3.51} $\pm$ 0.90\\
    \hline
    \end{tabular}
    \vspace{-1pt}
\end{table}
\begin{table}[t]
    \centering
    \caption{Real-robot Results on ``Grasping the Invisble''}
    \vspace{-2pt}
    \label{tab:real_table}
    \begin{tabular}{c|c|c}
    \hline
    Method & Task Success Rate ($\%$) & Number of Total Motions\\
    \hline
    VPG & 32.5 & 14.4\\
    Mask-VPG & 67.5 & 11.6\\
    Ours & \textbf{85.0} & \textbf{9.8}\\
    \hline
    \end{tabular}
    \vspace{-5pt}
\end{table}

We evaluate the methods on eight challenging test cases with adversarial structures shown in Fig. \ref{fig:structures}. In each test case, the maximum number of motions is 5, and the robot is tasked to grasp the target in clutter. We execute 30 runs on each test case and report the task success rate and the number of motions in Fig. \ref{fig:bar_coordinate_trail}. Our approach \textbf{Coordinator} outperforms the other approaches in terms of both the task success rate and the number of motions. Overall \textbf{Coordinator} achieves $87.5\%$ task success rate in the 8 challenging arrangements; \textbf{RAND} shows about $18\%$ chance of task success rate; the performance of \textbf{Mask-VPG} and \textbf{Border-Heuristic} are similar, while \textbf{Mask-VPG} shows slightly superior performance. \textbf{Mask-VPG} is originally designed for target-agnostic tasks, and thus it lacks the reasoning about the target and surrounding objects. Though \textbf{Border-Heuristic} has the same input with \textbf{Coordinator}, the hard-coded heuristic limits the coordination of target-oriented pushing and grasping. The comparison between \textbf{Border-Heuristic} and \textbf{Coordinator} shows the effectiveness of $\pi_c$ for coordination. As shown in Table \ref{tab:average_coordination}, \textbf{Coordinator} increases by more than $17\%$ in terms of the task success rate and requires about $0.6$ fewer average number of motions compared with \textbf{Mask-VPG} and \textbf{Border-Heuristic}.

\begin{figure}[t]
    \centering
    \begin{subfigure}{0.11\textwidth}
        \includegraphics[width=\textwidth]{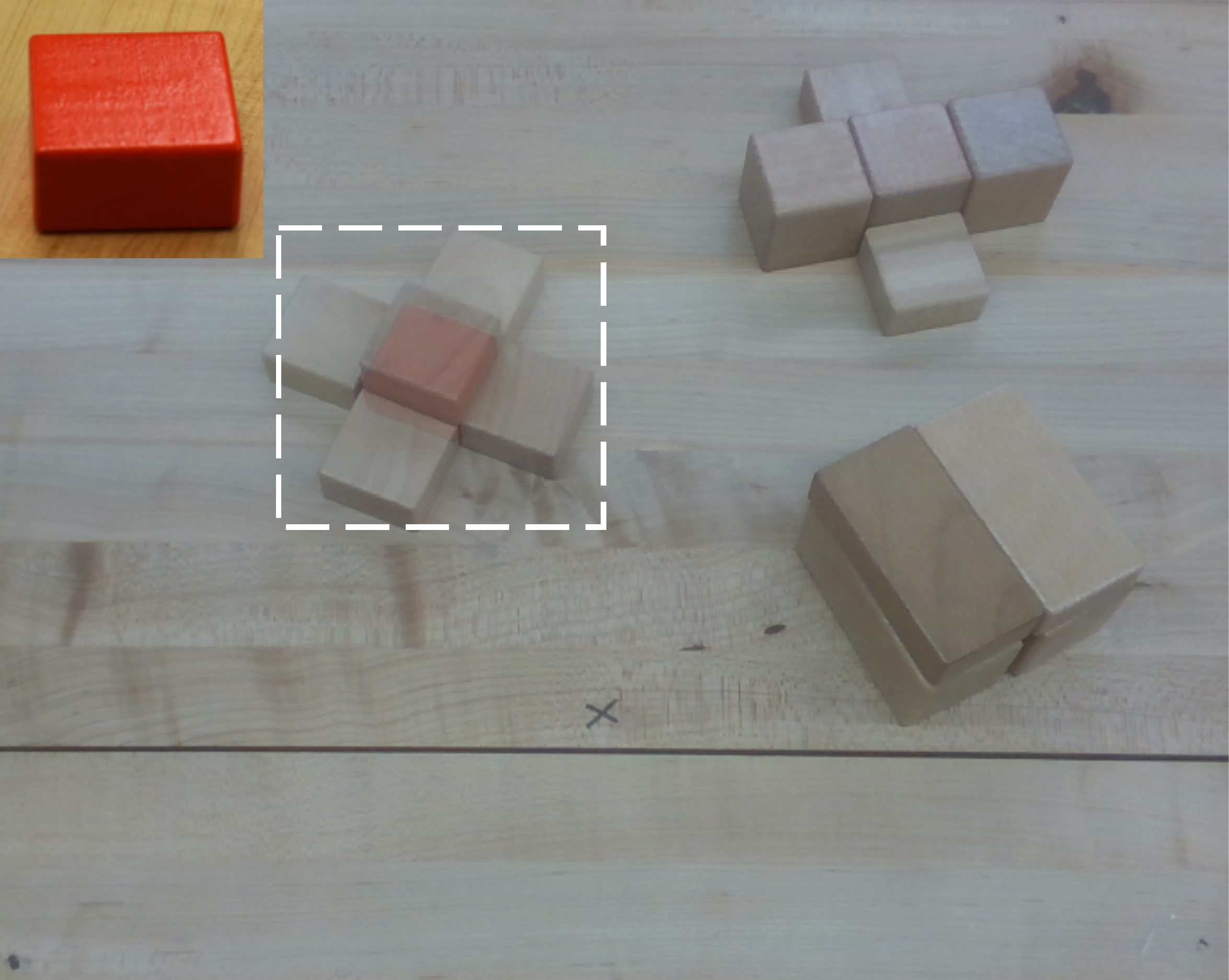}%
        \vspace{-5pt}
        \caption*{red cuboid}
    \end{subfigure}
    \begin{subfigure}{0.11\textwidth}
        \includegraphics[width=\textwidth]{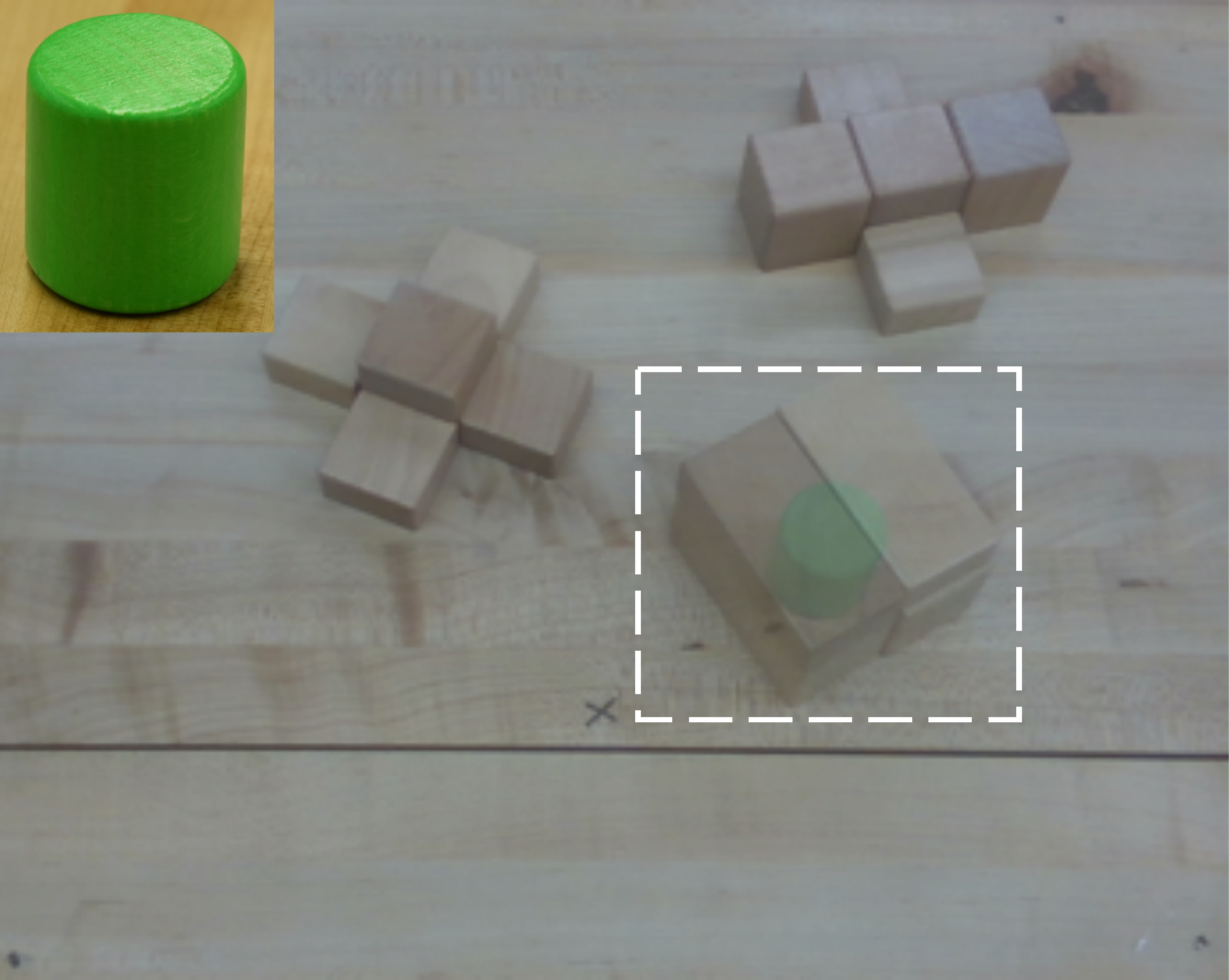}%
        \vspace{-5pt}
        \caption*{green cylinder}
    \end{subfigure}
    \begin{subfigure}{0.11\textwidth}
        \includegraphics[width=\textwidth]{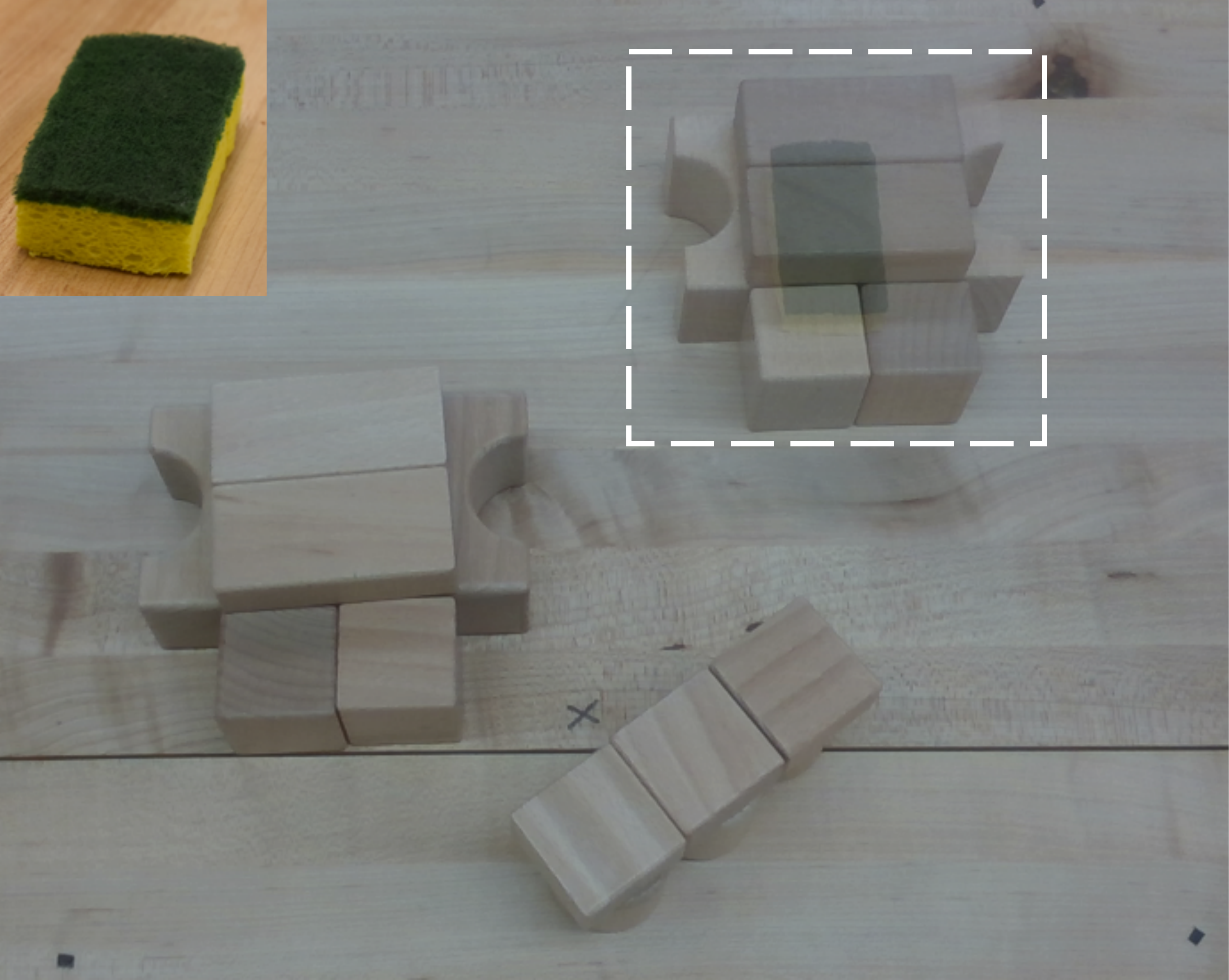}%
        \vspace{-5pt}
        \caption*{sponge}
    \end{subfigure}
    \begin{subfigure}{0.11\textwidth}
        \includegraphics[width=\textwidth]{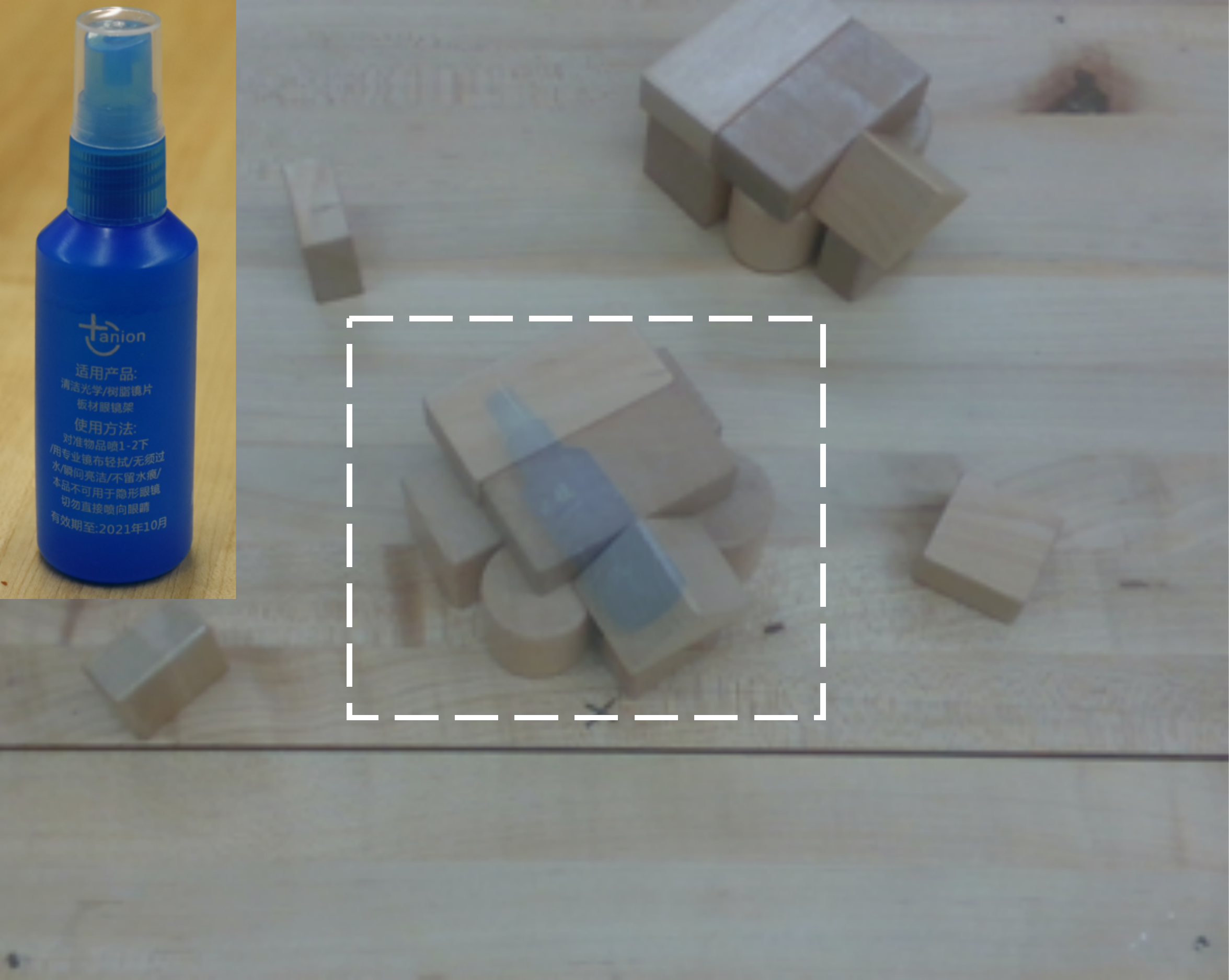}%
        \vspace{-5pt}
        \caption*{spray bottle}
    \end{subfigure}
    \vspace{-3pt}
    \caption{\textbf{Test cases on the real robot.} The invisible target is either a toy block or a novel object never seen in training.}
    \vspace{-5pt}
    \label{fig:real-test}
\end{figure}

\subsection{Real-robot Experiments}
We evaluate our system on the ``grasping the invisible'' problem with a Franka EMIKA Panda robot using the model trained in simulation\footnote{We don't fine-tune the model on the real robot. To collect data in the real world, the robot could hold a grasped object up to the camera and use an additional classifier to check grasping results \cite{fang2018multi}.}. For the comparison baselines, we use the available model trained with a robot system in the real world \cite{zeng2018learning}. Over 4 test cases, as shown in Fig. \ref{fig:real-test}, our approach and two baselines, \textbf{VPG} and \textbf{Mask-VPG}, are tested. Successful target grasping is manually checked for \textbf{VPG} as it is target-agnostic. When searching for the target, \textbf{Mask-VPG} works as \textbf{VPG} (i.e., no mask post-processing) except that only pushing is enabled for a fair comparison.

We run on each test case for 10 runs, and the maximum number of motions for each run is 15. The robot is tasked to find the initially invisible target as well as grasp it in challenging clutter. Table \ref{tab:real_table} reports the task success rate and the number of total motions of the three methods. Overall, our approach trained in simulation outperforms the domain-adapted baselines and achieves a task success rate of $85\%$. The results show our system is capable of generalizing to new environments, sets of objects of a similar shape to training objects, and adversarial arrangements. \textbf{VPG} shows a task success rate of only $32.5\%$ with a high average number of motions. Target-agnostic \textbf{VPG} tends to prioritize grasping easily graspable objects while the invisible target is buried in heavy clutter. \textbf{Mask-VPG} is advanced by reducing the action field by the target mask, and its task success rate improves to be $67.5\%$ with $11.6$ average number of motions. In contrast to \textbf{Mask-VPG}, our approach outperforms by $17.5\%$ in terms of the task success rate and requires about $1.8$ fewer average number of motions. Through experiments, we find that \textbf{Mask-VPG} tends to be less robust to noise in real settings and lacks the capability of superior coordination that our approach demonstrates.

\section{CONCLUSIONS}

In this work, we presented the ``grasping the invisible'' problem and proposed a deep learning approach in a critic-policy format. The learning models of our approach were trained by self-supervision in simulation. We evaluate the system in both simulated and real settings. Our approach shows a $93\%$ and $87\%$ task success rate on the two subtasks in simulation and an $85\%$ task success rate in the real robot experiments, which outperforms the other compared approaches by large margins. The evaluation results with the real robot show the generalization capability of our approach. The learned model in simulation was reliably transferred in the real setting and even generalizes to novel objects. 



\bibliographystyle{IEEEtran}
\bibliography{IEEEabrv,IEEEexample}

\begin{thebibliography}{10}
\providecommand{\url}[1]{#1}
\csname url@samestyle\endcsname
\providecommand{\newblock}{\relax}
\providecommand{\bibinfo}[2]{#2}
\providecommand{\BIBentrySTDinterwordspacing}{\spaceskip=0pt\relax}
\providecommand{\BIBentryALTinterwordstretchfactor}{4}
\providecommand{\BIBentryALTinterwordspacing}{\spaceskip=\fontdimen2\font plus
\BIBentryALTinterwordstretchfactor\fontdimen3\font minus
  \fontdimen4\font\relax}
\providecommand{\BIBforeignlanguage}[2]{{%
\expandafter\ifx\csname l@#1\endcsname\relax
\typeout{** WARNING: IEEEtran.bst: No hyphenation pattern has been}%
\typeout{** loaded for the language `#1'. Using the pattern for}%
\typeout{** the default language instead.}%
\else
\language=\csname l@#1\endcsname
\fi
#2}}
\providecommand{\BIBdecl}{\relax}
\BIBdecl

\bibitem{eitel17isrr}
A.~Eitel, N.~Hauff, and W.~Burgard, ``Learning to singulate objects using a
  push proposal network,'' in \emph{Proc. of the International Symposium on
  Robotics Research (ISRR)}, 2017.

\bibitem{choi2018learning}
C.~Choi, W.~Schwarting, J.~DelPreto, and D.~Rus, ``Learning object grasping for
  soft robot hands,'' \emph{IEEE Robotics and Automation Letters}, vol.~3,
  no.~3, pp. 2370--2377, 2018.

\bibitem{zeng2018learning}
A.~Zeng, S.~Song, S.~Welker, J.~Lee, A.~Rodriguez, and T.~Funkhouser,
  ``Learning synergies between pushing and grasping with self-supervised deep
  reinforcement learning,'' in \emph{2018 IEEE/RSJ International Conference on
  Intelligent Robots and Systems (IROS)}.\hskip 1em plus 0.5em minus
  0.4em\relax IEEE, 2018, pp. 4238--4245.

\bibitem{jang2018grasp2vec}
E.~Jang, C.~Devin, V.~Vanhoucke, and S.~Levine, ``Grasp2vec: Learning object
  representations from self-supervised grasping,'' in \emph{Conference on Robot
  Learning}, 2018, pp. 99--112.

\bibitem{fang2018multi}
K.~Fang, Y.~Bai, S.~Hinterstoisser, S.~Savarese, and M.~Kalakrishnan,
  ``Multi-task domain adaptation for deep learning of instance grasping from
  simulation,'' in \emph{2018 IEEE International Conference on Robotics and
  Automation (ICRA)}.\hskip 1em plus 0.5em minus 0.4em\relax IEEE, 2018, pp.
  3516--3523.

\bibitem{bohg2013data}
J.~Bohg, A.~Morales, T.~Asfour, and D.~Kragic, ``Data-driven grasp
  synthesis—a survey,'' \emph{IEEE Transactions on Robotics}, vol.~30, no.~2,
  pp. 289--309, 2013.

\bibitem{sahbani2012overview}
A.~Sahbani, S.~El-Khoury, and P.~Bidaud, ``An overview of 3d object grasp
  synthesis algorithms,'' \emph{Robotics and Autonomous Systems}, vol.~60,
  no.~3, pp. 326--336, 2012.

\bibitem{mahler2017dex}
J.~Mahler, J.~Liang, S.~Niyaz, M.~Laskey, R.~Doan, X.~Liu, J.~A. Ojea, and
  K.~Goldberg, ``Dex-net 2.0: Deep learning to plan robust grasps with
  synthetic point clouds and analytic grasp metrics,'' \emph{Robotics: Science
  and Systems (RSS)}, 2017.

\bibitem{kalashnikov2018scalable}
D.~Kalashnikov, A.~Irpan, P.~Pastor, J.~Ibarz, A.~Herzog, E.~Jang, D.~Quillen,
  E.~Holly, M.~Kalakrishnan, V.~Vanhoucke \emph{et~al.}, ``Scalable deep
  reinforcement learning for vision-based robotic manipulation,'' in
  \emph{Conference on Robot Learning}, 2018, pp. 651--673.

\bibitem{dogar2012planning}
M.~R. Dogar and S.~S. Srinivasa, ``A planning framework for non-prehensile
  manipulation under clutter and uncertainty,'' \emph{Autonomous Robots},
  vol.~33, no.~3, pp. 217--236, 2012.

\bibitem{cosgun2011push}
A.~Cosgun, T.~Hermans, V.~Emeli, and M.~Stilman, ``Push planning for object
  placement on cluttered table surfaces,'' in \emph{2011 IEEE/RSJ international
  conference on intelligent robots and systems}.\hskip 1em plus 0.5em minus
  0.4em\relax IEEE, 2011, pp. 4627--4632.

\bibitem{hermans2012guided}
T.~Hermans, J.~M. Rehg, and A.~Bobick, ``Guided pushing for object
  singulation,'' in \emph{2012 IEEE/RSJ International Conference on Intelligent
  Robots and Systems}.\hskip 1em plus 0.5em minus 0.4em\relax IEEE, 2012, pp.
  4783--4790.

\bibitem{chang2012interactive}
L.~Chang, J.~R. Smith, and D.~Fox, ``Interactive singulation of objects from a
  pile,'' in \emph{2012 IEEE International Conference on Robotics and
  Automation}.\hskip 1em plus 0.5em minus 0.4em\relax IEEE, 2012, pp.
  3875--3882.

\bibitem{danielczuk2018linear}
M.~Danielczuk, J.~Mahler, C.~Correa, and K.~Goldberg, ``Linear push policies to
  increase grasp access for robot bin picking,'' in \emph{2018 IEEE 14th
  International Conference on Automation Science and Engineering (CASE)}.\hskip
  1em plus 0.5em minus 0.4em\relax IEEE, 2018, pp. 1249--1256.

\bibitem{marios2019robust}
K.~Marios and M.~Sotiris, ``Robust object grasping in clutter via
  singulation,'' in \emph{2019 International Conference on Robotics and
  Automation (ICRA)}.\hskip 1em plus 0.5em minus 0.4em\relax IEEE, 2019, pp.
  1596--1600.

\bibitem{boularias2015learning}
A.~Boularias, J.~A. Bagnell, and A.~Stentz, ``Learning to manipulate unknown
  objects in clutter by reinforcement,'' in \emph{Twenty-Ninth AAAI Conference
  on Artificial Intelligence}, 2015.

\bibitem{jang2017end}
E.~Jang, S.~Vijayanarasimhan, P.~Pastor, J.~Ibarz, and S.~Levine, ``End-to-end
  learning of semantic grasping,'' in \emph{Conference on Robot Learning},
  2017, pp. 119--132.

\bibitem{danielczuk2019mechanical}
M.~{Danielczuk}, A.~{Kurenkov}, A.~{Balakrishna}, M.~{Matl}, D.~{Wang},
  R.~{Martín-Martín}, A.~{Garg}, S.~{Savarese}, and K.~{Goldberg},
  ``Mechanical search: Multi-step retrieval of a target object occluded by
  clutter,'' in \emph{2019 International Conference on Robotics and Automation
  (ICRA)}, May 2019, pp. 1614--1621.

\bibitem{ioffe2015batch}
S.~Ioffe and C.~Szegedy, ``Batch normalization: Accelerating deep network
  training by reducing internal covariate shift,'' in \emph{International
  Conference on Machine Learning}, 2015, pp. 448--456.

\bibitem{nair2010rectified}
V.~Nair and G.~E. Hinton, ``Rectified linear units improve restricted boltzmann
  machines,'' in \emph{Proceedings of the 27th international conference on
  machine learning (ICML-10)}, 2010, pp. 807--814.

\bibitem{badrinarayanan2017segnet}
V.~Badrinarayanan, A.~Kendall, and R.~Cipolla, ``Segnet: A deep convolutional
  encoder-decoder architecture for image segmentation,'' \emph{IEEE
  transactions on pattern analysis and machine intelligence}, vol.~39, no.~12,
  pp. 2481--2495, 2017.

\bibitem{he2016deep}
K.~He, X.~Zhang, S.~Ren, and J.~Sun, ``Deep residual learning for image
  recognition,'' in \emph{Proceedings of the IEEE conference on computer vision
  and pattern recognition}, 2016, pp. 770--778.

\bibitem{huang2017densely}
G.~Huang, Z.~Liu, L.~Van Der~Maaten, and K.~Q. Weinberger, ``Densely connected
  convolutional networks,'' in \emph{Proceedings of the IEEE conference on
  computer vision and pattern recognition}, 2017, pp. 4700--4708.

\bibitem{deng2009imagenet}
J.~Deng, W.~Dong, R.~Socher, L.-J. Li, K.~Li, and L.~Fei-Fei, ``Imagenet: A
  large-scale hierarchical image database,'' in \emph{2009 IEEE conference on
  computer vision and pattern recognition}.\hskip 1em plus 0.5em minus
  0.4em\relax Ieee, 2009, pp. 248--255.

\bibitem{pinto2017learning}
L.~Pinto and A.~Gupta, ``Learning to push by grasping: Using multiple tasks for
  effective learning,'' in \emph{2017 IEEE International Conference on Robotics
  and Automation (ICRA)}.\hskip 1em plus 0.5em minus 0.4em\relax IEEE, 2017,
  pp. 2161--2168.

\bibitem{zeng2018vpg}
\BIBentryALTinterwordspacing
A.~Zeng, ``visual-pushing-grasping,'' 2018. [Online]. Available:
  \url{https://github.com/andyzeng/visual-pushing-grasping}
\BIBentrySTDinterwordspacing

\bibitem{xiang2017posecnn}
Y.~Xiang, T.~Schmidt, V.~Narayanan, and D.~Fox, ``Posecnn: A convolutional
  neural network for 6d object pose estimation in cluttered scenes,''
  \emph{Robotics: Science and Systems (RSS)}, 2018.

\bibitem{periyasamy2018robust}
A.~S. Periyasamy, M.~Schwarz, and S.~Behnke, ``Robust 6d object pose estimation
  in cluttered scenes using semantic segmentation and pose regression
  networks,'' in \emph{2018 IEEE/RSJ International Conference on Intelligent
  Robots and Systems (IROS)}.\hskip 1em plus 0.5em minus 0.4em\relax IEEE,
  2018, pp. 6660--6666.

\bibitem{wang2019densefusion}
C.~Wang, D.~Xu, Y.~Zhu, R.~Mart{\'\i}n-Mart{\'\i}n, C.~Lu, L.~Fei-Fei, and
  S.~Savarese, ``Densefusion: 6d object pose estimation by iterative dense
  fusion,'' in \emph{Proceedings of the IEEE Conference on Computer Vision and
  Pattern Recognition}, 2019, pp. 3343--3352.

\bibitem{nekrasov2018light}
V.~Nekrasov, C.~Shen, and I.~Reid, ``Light-weight refinenet for real-time
  semantic segmentation,'' \emph{arXiv preprint arXiv:1810.03272}, 2018.

\bibitem{schaul2015prioritized}
T.~Schaul, J.~Quan, I.~Antonoglou, and D.~Silver, ``Prioritized experience
  replay,'' \emph{arXiv preprint arXiv:1511.05952}, 2015.

\bibitem{andrychowicz2017hindsight}
M.~Andrychowicz, F.~Wolski, A.~Ray, J.~Schneider, R.~Fong, P.~Welinder,
  B.~McGrew, J.~Tobin, O.~P. Abbeel, and W.~Zaremba, ``Hindsight experience
  replay,'' in \emph{Advances in Neural Information Processing Systems}, 2017,
  pp. 5048--5058.

\end{thebibliography}

\newpage
\section{APPENDIX}
\setcounter{subsection}{0}
\subsection{Action Policy of Mechanical Search}\label{appen:mech_search}
We summarize the action policy of Mechanical Search \cite{danielczuk2019mechanical} for comparison purposes. The action policy of \cite{danielczuk2019mechanical} takes as input the masks of objects in the bin and selects the action to execute. The policy is composed of an action selection method and an action execution criterion. The action selection method (e.g., largest visible object first) is responsible for keeping a priority list. And the action execution criterion iterates the objects in the list to make the action decision. The action execution criterion can be summarized as Algorithm \ref{alg:mech_search}. For every object in the priority list, Algorithm \ref{alg:mech_search} selects an action or reports a failure until the target is retrieved.

The formulation of a POMDP is given in \cite{danielczuk2019mechanical} for the target retrieval problem. However, due to the computation difficulty, Algorithm \ref{alg:mech_search} is proposed to apply action decision making heuristically instead of solving the POMDP problem in the reinforcement learning framework.

In addition, pushing is of low priority in Algorithm \ref{alg:mech_search}, i.e., pushing is only considered without a satisfactory grasping being found. In the experiments, \cite{danielczuk2019mechanical} shows pushing makes up only $5\%$ of executed actions. More effective
push primitives are listed as one of the future work in \cite{danielczuk2019mechanical}.

\begin{algorithm}
\caption{Action Execution Criterion}
\textbf{Input}: object mask $o$\\
\textbf{Output}: selected action $a$ or termination\\
\textbf{Notations}: threshold $t$, action $a$, action quality $q$ 
\begin{algorithmic}[1]
\If{$o$ is the search target or pushing is disabled}
\State $t_\text{grasp} \gets t_\text{low}$
\Else
\State $t_\text{grasp} \gets t_\text{high}$
\EndIf
\State $a,q \gets \text{GraspingModules}(o)$
\If{$q > t_\text{grasp}$}
\State execute grasp action a
\Else
\If{pushing enabled}
\State $a,q \gets \text{PushingModule}(o)$
\If{valid $a$ not found or pushed three times}
\State termination with a failure
\EndIf
\Else
\State termination with a failure
\EndIf
\EndIf
\end{algorithmic}
\label{alg:mech_search}
\end{algorithm}

\subsection{Pre-trained Perception
Module}\label{appen:seg}
\begin{figure}[t]
    \centering
    \begin{subfigure}{0.21\textwidth}
        \centering
        \includegraphics[width=\textwidth]{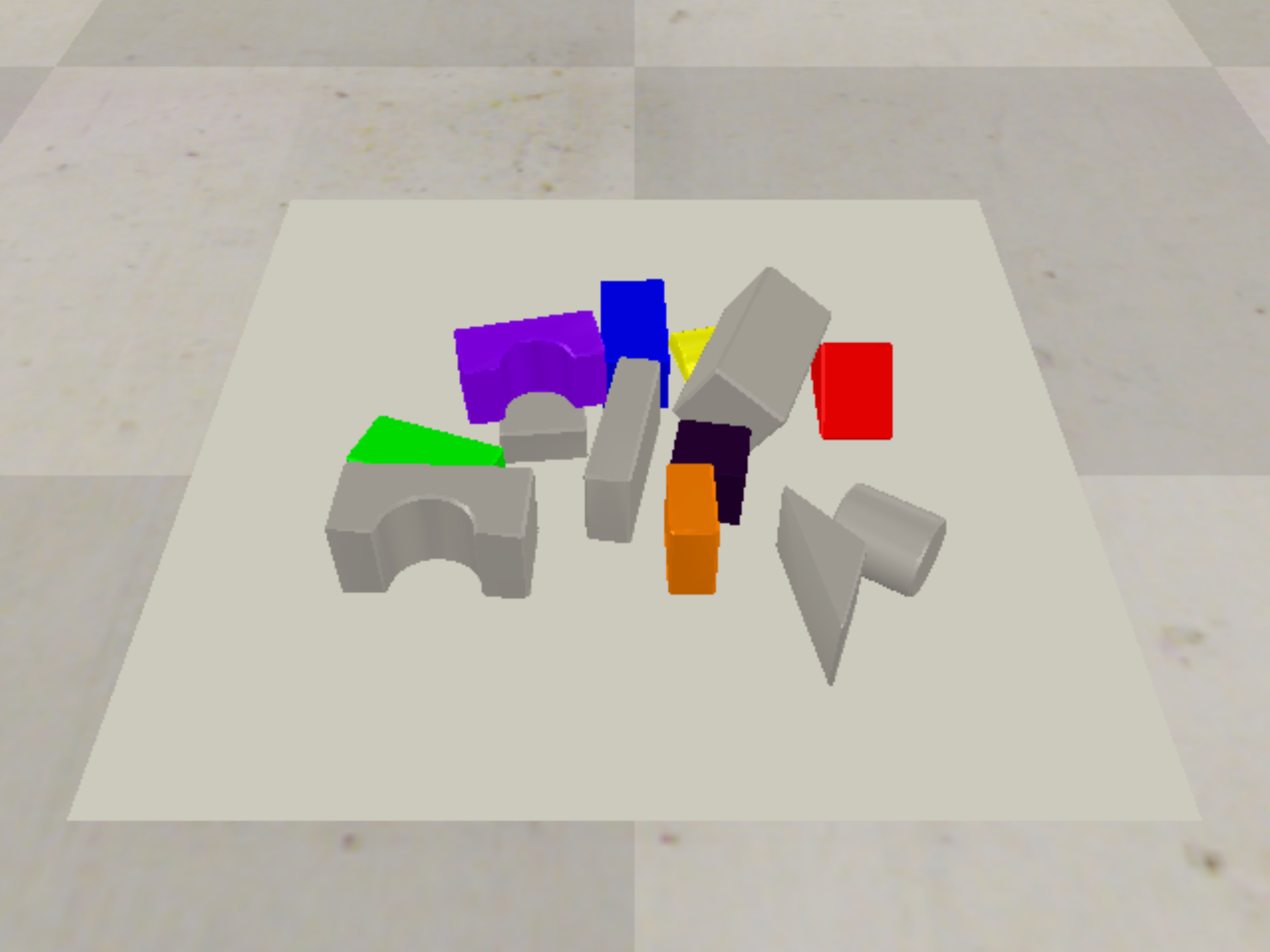}%
        \vspace{-3pt}
        \caption*{Cluttered scenes}
    \end{subfigure}
    \hfill
    \begin{subfigure}{0.21\textwidth}  
        \centering
        \includegraphics[width=\textwidth]{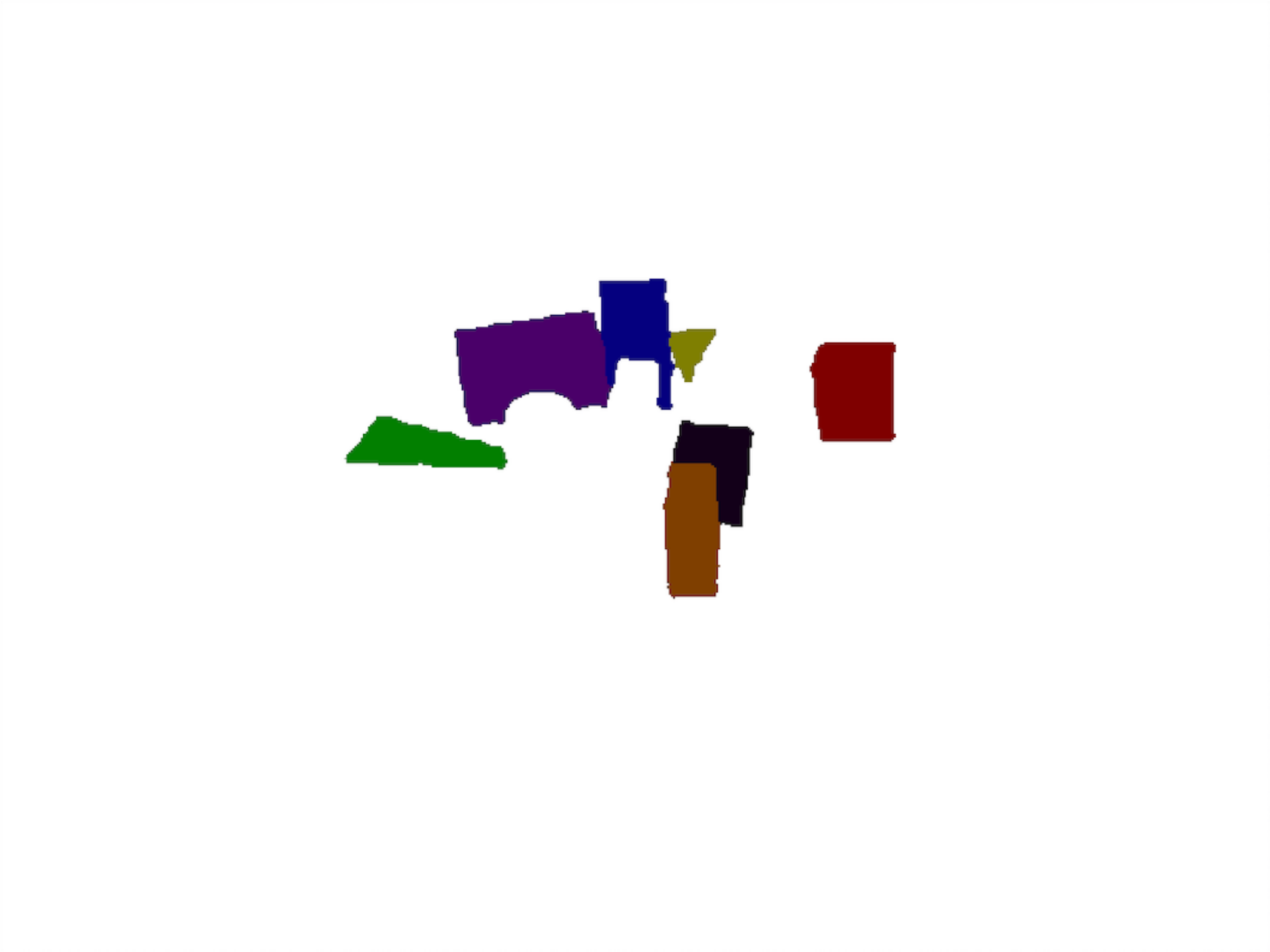}%
        \vspace{-3pt}
        \caption*{Segmentation results}
    \end{subfigure}
    \vspace{-5pt}
    \caption{\textbf{Example of semantic image segmentation.} The background is visualized in white color for best visualization.}
    \vspace{-8pt}
    \label{fig:seg}
\end{figure}
Semantic segmentation is widely used in 6D object pose estimation in cluttered scenes \cite{xiang2017posecnn}, \cite{periyasamy2018robust}, \cite{wang2019densefusion} and proves to be very robust to occlusions between objects. We choose Light-Weight RefineNet \cite{nekrasov2018light} as our perception module and pre-train it on the augmented data. Inspired by \cite{periyasamy2018robust}, we synthetically generated the training dataset covering all target candidates, rich pose variations of the objects, and occlusions with a handful of labeled data. The pre-trained model robustly segments the cluttered scenes, as shown in Fig \ref{fig:seg}.

\subsection{Additional Training Details}\label{appen:train}
We save states, actions and rewards into experience replay buffer $B_c$ and train the critic with prioritized experience replay \cite{schaul2015prioritized}. To deal with sparse rewards for target-oriented grasping (especially at early stages), the hindsight experience replay technique \cite{andrychowicz2017hindsight} is used. If a non-target object is grasped at time $t$, we save executed action $a_t$, states, mask of the grasped object $m_{t}^{'}$ and posthoc labeled reward $R_{a_t}^{'}((c_t,d_t,m_{t}^{'}),s_{t+1})=1$ for further experience replay training. To train action classifier $f_a$, we save Q values and domain knowledge, and the generated label into bounded buffer $B_{\pi}$ (a cyclic buffer of bounded size). The training process is summarized in Algorithm \ref{alg:training}.

\begin{algorithm}
\caption{Training Algorithm}
\textbf{Objects}: critic $C$, policy $\pi_c$ (with classifier $f_a$)\\
\textbf{Initialize} experience replay buffer $B_c$ for $C$\\ \textbf{Initialize} bounded buffer $B_{\pi}$ for $\pi_c$\\
\textbf{Notations}: iteration $i$, $\epsilon$-greedy policy $\pi_\epsilon$, target $T$, image $I$, mask $M$, state $s$, push or grasp network $\phi$, maximum Q value $q$, domain knowledge $D=(r_b,n_b,c_g)$
\begin{algorithmic}[1]
\State $i \gets 0$
\While{True}
\State reset the simulation and randomly drop objects
\State randomly select a visible target $T$
\While{$T$ not grasped}
\State $i \gets i+1$
\State $M$ $\gets$ \texttt{ObjectSegmentation}($I$, $T$)
\State $s_t$ $\gets$ \texttt{HeightmapProjection}($I$,$M$)
\State $Q_p \gets \phi_p(s_t)$, $Q_g \gets \phi_g(s_t)$
\If{$i < 1000$}
\State execute action $a_t \gets \pi_{\epsilon}(Q_p,Q_g)$
\Else
\State execute action $a_t \gets \pi_{c}(Q_p,Q_g,D)$
\State label $\bar{y}$ according to grasping results
\State save $(q_p,q_g,D)$ and $\bar{y}$ into $B_{\pi}$
\State sample a batch from $B_{\pi}$ to train $f_a$
\EndIf
\State $R_t \gets \texttt{RewardFunction}(\cdot)$
\State save $s_t$, $a_t$, $R_t$ into $B_c$
\If{non-target object grasped}
\State save extra mask $m_{t}^{'}$ and reward $R_t^{'}$ into $B_c$
\EndIf
\State sample data from $B_c$ to train $C$
\EndWhile
\EndWhile
\end{algorithmic}
\label{alg:training}
\end{algorithm}

\subsection{Additional Details of Pushing Failure Likelihood}\label{appen:push_failures}
We first construct a 2D Gaussian function $G$ with the peak at the center of the workspace. The standard deviation of $G$ is set to cover the pushing length.  Then we apply a linear transform on $G$ to get function $G^{'}$
\begin{align}
    G^{'} = 1-\frac{(1-s)G}{\max(G)}
\end{align}
where $s$ is the scale parameter and is set to be $0.75$ in our experiments. We shift $G^{'}$ to the location of a failed action to reduce the probability around the failed locations. The shifting is repeated for the three most recent locations of failed pushes, and we multiply the shifted functions to construct pushing failure likelihood $F_p$. In the implementation, we use a deque to store the most recent shifted functions for performance.

\subsection{Additional Details of Border-Heuristic}\label{appen:baseline}
\textbf{Border-Heuristic} determines whether to push or grasp by a heuristic. Once the decision is made, pushing or grasping is executed by target-oriented motion critic. We include Border-Heuristic as a baseline to show how much a learned classifier (in our coordination policy) outperforms a hand-engineered heuristic. \textbf{Border-Heuristic} is an exploration-based heuristic, and the pushing exploration rate is defined as
\begin{align}
    f_\epsilon(\cdot) = \min(1.0, \max(0.0, 0.5-\alpha\frac{q_g-q_p}{(c_g+1)(r_b+n_b)}))
\end{align}
where $\alpha=1e^{-2}$ is the hand-engineered parameter in our experiments. The algorithm of \textbf{Border-Heuristic} is summarized in Algorithm \ref{alg:border-heuristic}.
\begin{algorithm}
\caption{Border-Heuristic}
\textbf{Input}: push and grasp Q maps, target border occupancy ratio $r_b$, target border occupancy norm $n_b$, the number of consecutive grasping failures $c_g$\\
\textbf{Output}: motion type $y \in \{p,g\}$ and action decision $a_t$
\begin{algorithmic}[1]
\State $q_p \gets \max Q_p$, $q_g \gets \max Q_g$
\State $\epsilon_p = f_\epsilon(q_p,q_g,r_b,n_b,c_g)$
\If{random$() < \epsilon_p$}
\State $y \gets p$
\EndIf
\If{$y = g$ and $c_g \neq 0$}
\If{$c_g<2$}
\If{random$() < r_b$}
\State $y \gets p$
\EndIf
\Else
\State $y \gets p$
\EndIf
\EndIf
\State $a_t \gets \underset{a}\max \, Q_y$
\end{algorithmic}
\label{alg:border-heuristic}
\end{algorithm}
\begin{figure}[t]
    \centering
    \begin{subfigure}{0.48\textwidth}
        \centering
        \includegraphics[width=\textwidth]{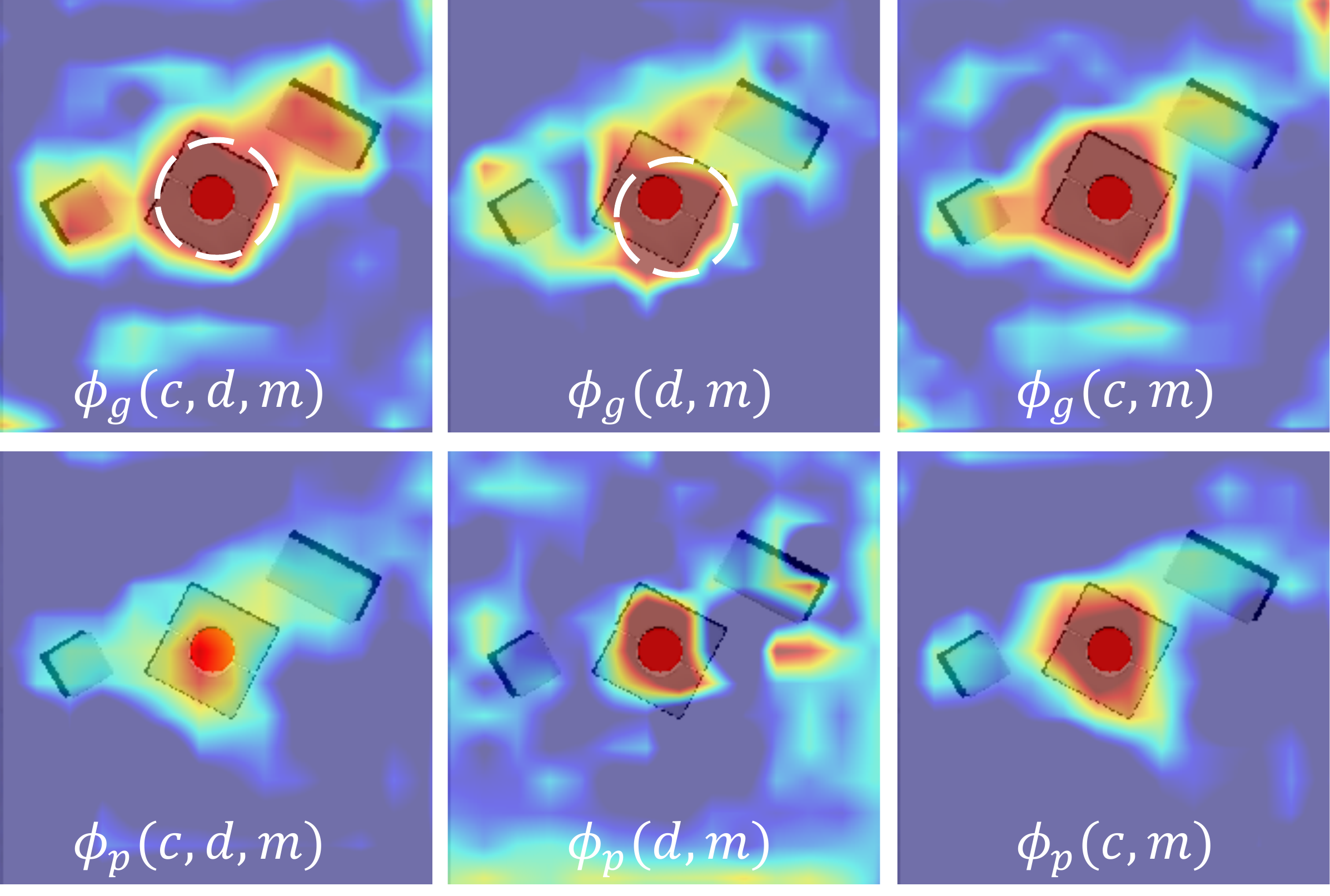}%
        \vspace{-3pt}
        \caption{Q maps in simulation}
        \vspace{4pt}
    \end{subfigure}
    \begin{subfigure}{0.48\textwidth}  
        \centering
        \includegraphics[width=\textwidth]{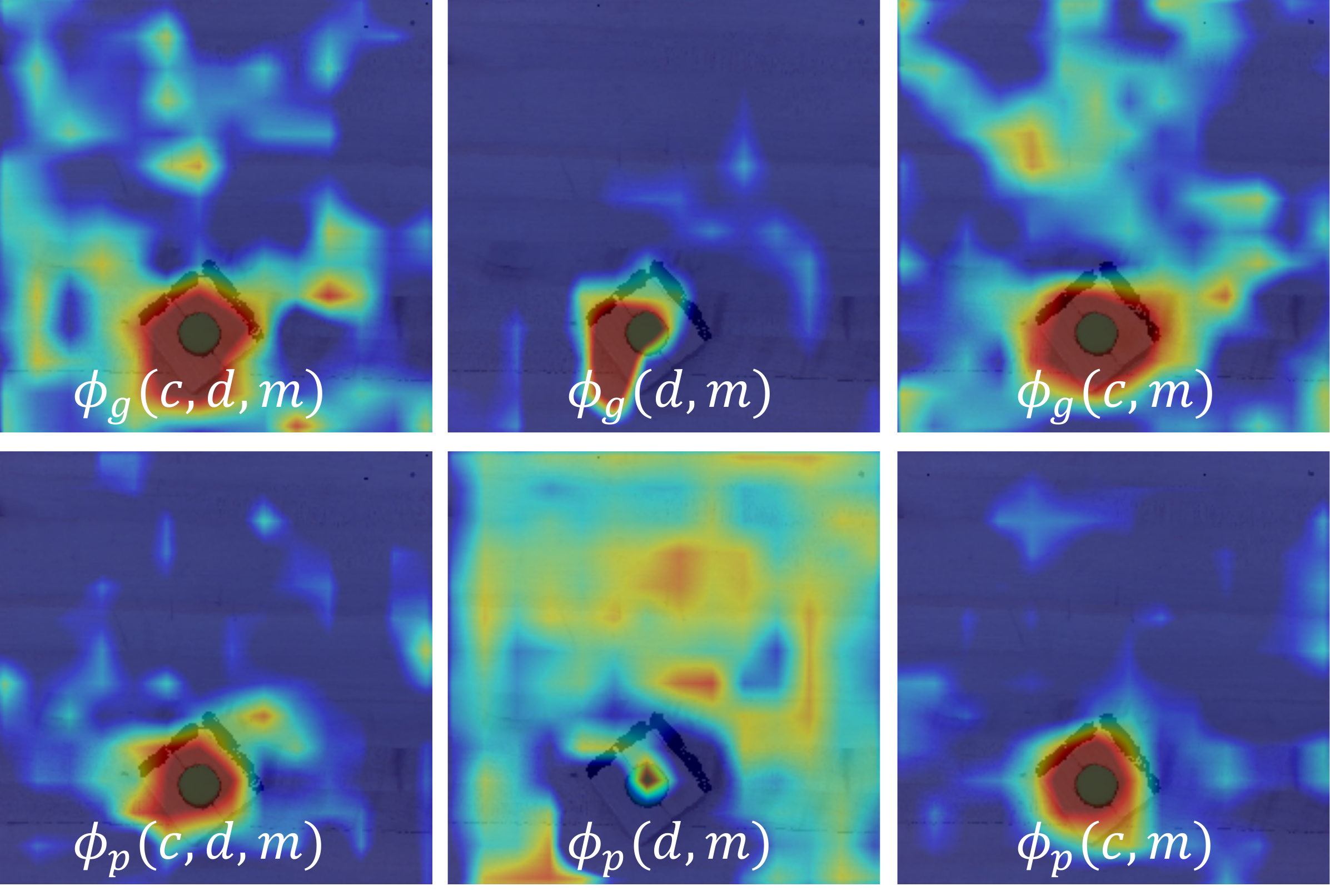}%
        \vspace{-3pt}
        \caption{Q maps in the real world}
    \end{subfigure}
    \vspace{-5pt}
    \caption{\textbf{Testing with different inputs and in different domains.} The maps are generated by our pre-trained model.}
    \vspace{-8pt}
    \label{fig:add_testing}
\end{figure}

\subsection{Additional Testing}\label{appen:testing}
We test our model with different inputs and in different domains. We visualize the output of the motion critic (composed of push network $\phi_p$ and grasp network $\phi_g$ ) with different combinations of color, depth, and mask input (denoted as $c$, $d$ and $m$) in Fig. \ref{fig:add_testing}. The Q maps are produced by one pre-trained model, and the corresponding CNN parameters are set to zero for the disabled channel. 

We observe that both color and depth channel affect the output. There are nontrivial changes if we disable either one of the channels. For example, we notice that, in simulation, the Q map shifts away from the target if we disable the color channel, as shown with the white dashed circle in Fig. \ref{fig:add_testing}.

Our model generalizes to the real world since $\phi(c, d, m)$ shares some similar patterns when transferring from simulation to the real world. For example, $\phi_g(c, d, m)$ and $\phi_p(c, d, m)$ have higher Q values around the target in both simulation and the real world.

\subsection{Additional Training}
To further investigate the importance of the color channel in our approach, we train a depth+mask variant of our approach. We make minimal modifications on the motion critic to accept only the depth and the target mask while keeping all other parts unchanged. We evaluate \textbf{Depth+Mask} and report the results in Table \ref{tab:depth+mask}. Our approach outperforms \textbf{Depth+Mask}.
\begin{figure}[t]
    \centering
    \begin{subfigure}{0.48\textwidth}
        \centering
        \includegraphics[width=\textwidth]{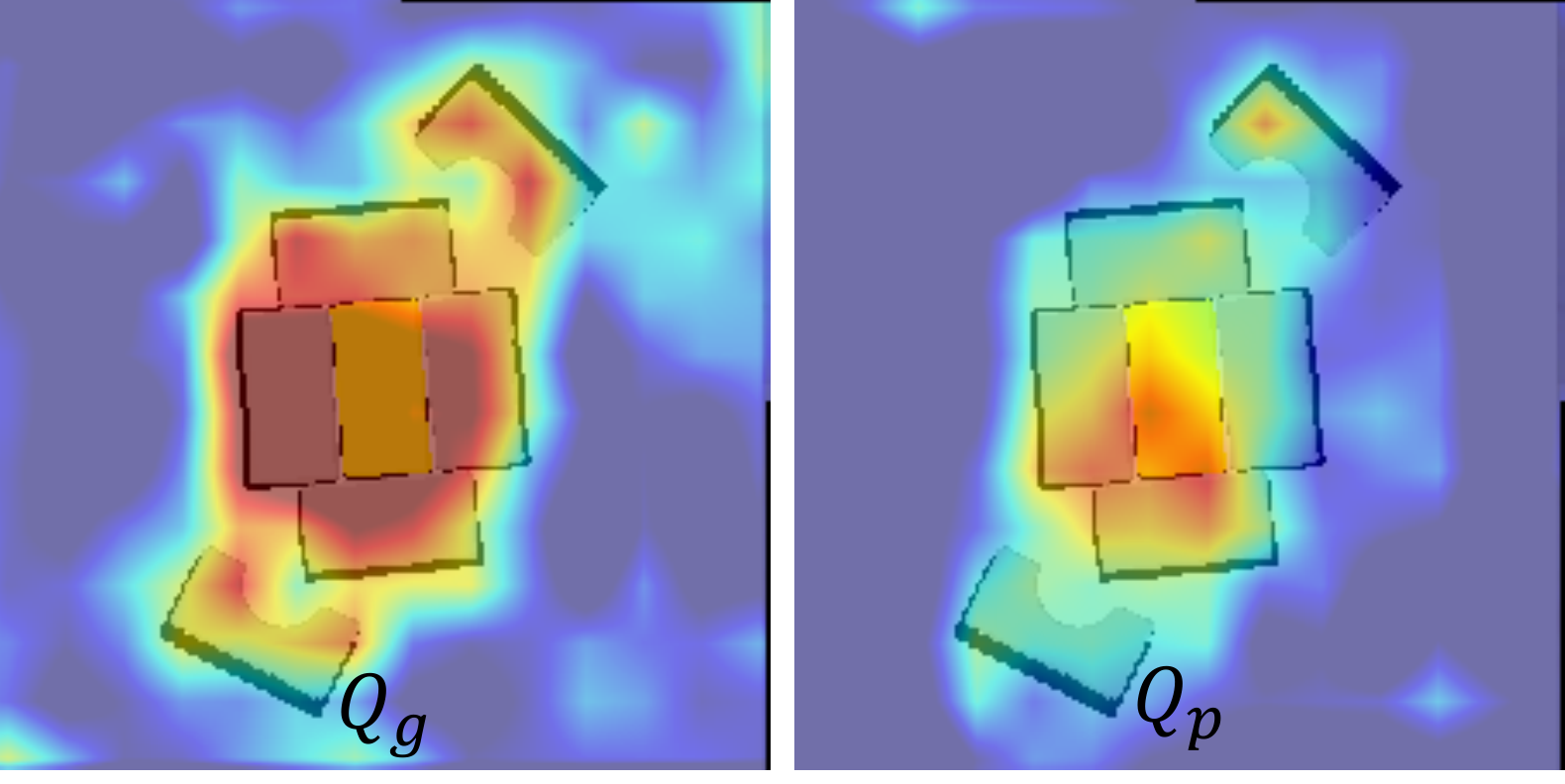}%
        \vspace{-3pt}
        \caption{Q maps of Ours (color+depth+mask)}
        \vspace{4pt}
    \end{subfigure}
    \begin{subfigure}{0.48\textwidth}  
        \centering
        \includegraphics[width=\textwidth]{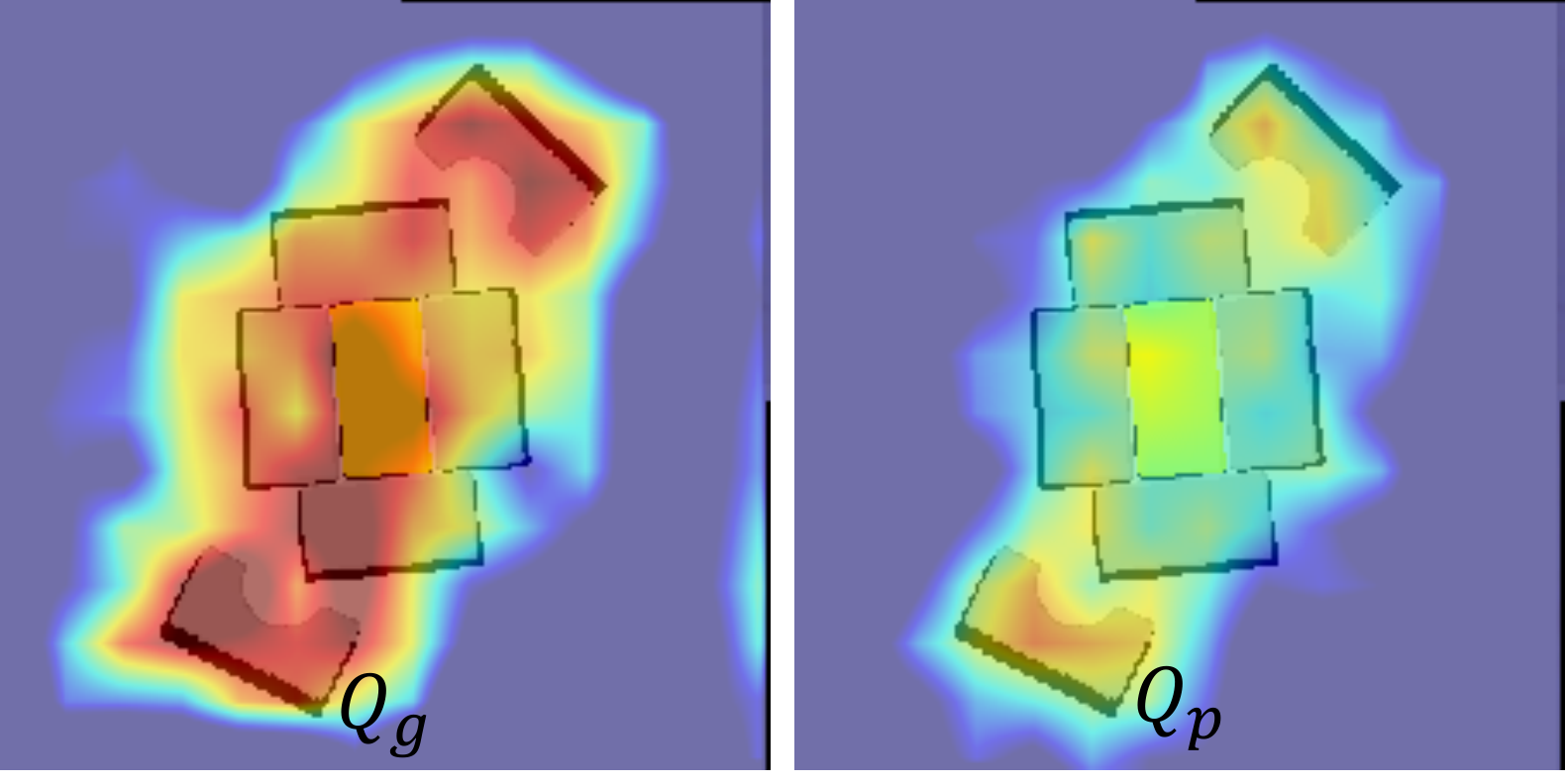}%
        \vspace{-3pt}
        \caption{Q maps of Depth+Mask}
    \end{subfigure}
    \vspace{-5pt}
    \caption{\textbf{Comparison of Q maps.} The push and grasp maps are for one test case.}
    \vspace{-8pt}
    \label{fig:add_training}
\end{figure}
\begin{table}[h!]
    \centering
    \caption{Average Performance of depth+mask}
    \label{tab:depth+mask}
    \begin{tabular}{c|c|c}
    \hline
    Method & Success Rate ($\%$) & Number of Motions\\
    \hline
    Depth+Mask in exploration & 87.0 & 3.30 $\pm$ 1.34\\
    Ours in exploration & \textbf{93.5} & \textbf{2.70} $\pm$ 1.28\\
    \hline
    \hline
    Depth+Mask in coordination & 78.7 & 4.06 $\pm$ 0.29\\
    \hline
    Ours in coordination & \textbf{87.5} & \textbf{3.51} $\pm$ 0.90\\
    \hline
    \end{tabular}
\end{table}

We conjecture that the color input helps the networks to distinguish the target from surrounding objects, especially in generalization. As shown in Fig. \ref{fig:add_training}, when we test \textbf{Depth+Mask} on the challenging arrangements never seen during training, the Q maps shift from the target compared to the Q maps of our approach. This observation is consistent with Appendix \ref{appen:testing}.

\end{document}